\documentclass{article}

    \PassOptionsToPackage{numbers, compress}{natbib}
 \usepackage[preprint]{neurips_2026}

\usepackage[utf8]{inputenc} 
\usepackage[T1]{fontenc}    
\usepackage{hyperref}       
\usepackage{url}            
\usepackage{booktabs}       
\usepackage{amsfonts}       
\usepackage{nicefrac}       
\usepackage{microtype}      
\usepackage{xcolor}         
\usepackage{amsmath}
\usepackage{microtype}
\usepackage{graphicx}
\usepackage{subcaption}

\usepackage{hyperref}
\usepackage{booktabs}   
\usepackage{multirow}   
\usepackage{array}  
\usepackage{tabularx}
\usepackage{lipsum}
\usepackage{tcolorbox}
\tcbuselibrary{raster}
\tcbuselibrary{breakable}

\usepackage{tikz}
\usepackage{lipsum}

\usepackage{amsmath}
\usepackage{amssymb}
\usepackage{mathtools}
\usepackage{amsthm}
\usepackage{tcolorbox}
\usepackage{longtable}
\usepackage{booktabs}
\usepackage{bm}
\usepackage{wrapfig}
\usepackage[table]{xcolor}
\usepackage{titletoc}
\definecolor{oursrow}{RGB}{230, 244, 234}  

\usepackage{pgfplots}
\pgfplotsset{compat=1.18}
\usepgfplotslibrary{groupplots}
\usetikzlibrary{positioning}
\usepackage{bbm}
\usepackage{xcolor}
\usepackage{colortbl}
\pgfplotsset{
  empty legend/.style={
    legend image code/.code={}
  }
}

\usepackage{bm} 

\newcommand{\best}[1]{\bm{#1}}
\newcommand{\second}[1]{\underline{#1}}

\definecolor{aswincolor}{RGB}{214,39,40}

\title{Mid-Training with Self-Generated Data Improves Reinforcement Learning in Language Models}

\newcommand{\authname}[2]{{\bfseries #1}\textsuperscript{\normalfont #2}}

\newcommand{\blfootnote}[1]{%
  \begingroup
  \renewcommand{\thefootnote}{}%
  \footnotetext{#1}%
  \endgroup
}

\author{%
\begin{tabular}{@{}c@{}}
\authname{Aswin RRV}{1}
\quad
\authname{Jacob Dineen}{1}
\quad
\authname{Divij Handa}{1}
\quad
\authname{Mihir Parmar}{2}
\\[0.35em]
\authname{Ben Zhou}{1}
\quad
\authname{Swaroop Mishra}{3$\diamond$}
\quad
\authname{Chitta Baral}{1}
\\[1.0em]
{\normalfont \textsuperscript{1}Arizona State University}
\quad
{\normalfont \textsuperscript{2}Google Cloud AI Research}
\quad
{\normalfont \textsuperscript{3}Google DeepMind}
\end{tabular}
}

\begin{document}

\maketitle

\blfootnote{%
Correspondence to: \texttt{aravik13@asu.edu}
\quad
\textsuperscript{$\diamond$}Work done prior to joining Google DeepMind.
}

\begin{abstract}

The effectiveness of Reinforcement Learning (RL) in Large Language Models (LLMs) depends on the nature and diversity of the data used before and during RL. In particular, reasoning problems can often be approached in multiple ways that rely on different forms of reasoning, and exposure to only a limited range of such approaches in the training data may limit the effectiveness of RL. Motivated by this, we investigate using diverse self-generated data during mid-training as an intermediate step before RL training. Specifically, we adopt a bootstrapped data-generation framework guided by George Pólya’s problem-solving approaches for generating multiple variants of correct answers for each question in the training data, and then perform fine-tuning. We first provide a theoretical perspective on how mid-training on such data improves RL and explain how policy-gradient updates can incentivize combining multiple approaches. We then empirically demonstrate that RL-trained models initialized with our mid-training data achieve consistent improvements across various mathematical reasoning benchmarks and other OOD tasks like code generation and narrative reasoning. Overall, our investigative study shows that a language model learning multiple problem-solving approaches, through self-generated data helps subsequent RL.

\end{abstract}

\section{Introduction}
Reinforcement Learning (RL) has recently emerged as an effective technique for improving the performance of LLMs, particularly on complex reasoning tasks. By encouraging longer reasoning chains, RL enables performance gains through inference-time scaling. Such reasoning chains often display behaviors, including multi-step reasoning, self-verification, and self-correction, that resemble human-like problem-solving~\citep{openai_o1,deepseek-r1, rlvr_lambert}. However, recent works argue that RL with simple, verifiable rewards (RLVR) primarily performs distributional sharpening \citep{zhou2025breaking,he2025rewarding}. It selectively amplifies behaviors already present in the base model rather than inducing genuinely new capabilities \citep{yue2025reinforcement}. In some cases, RLVR can even degrade performance under pass@k evaluation absent intervention \citep{entropymech,forking}. This highlights that RL effectiveness is closely tied to what the model learned prior to RL. 

Recent studies show that RL struggles to elicit self-reflective behaviors in Llama-3.2 family models that lack strong priors \citep{cognitive_behaviors,think_tuning}. Furthermore, \citet{zhang2025interplaypretrainingmidtrainingrl} demonstrate that capability gains beyond the base model are achievable only under specific conditions on both the task and the training data. For reasoning tasks, this observation is particularly relevant. A single problem can often be solved in multiple valid ways, relying on different forms of reasoning. When a model lacks exposure to diverse approaches, subsequent RL training may struggle to improve its performance. Together, these observations highlight the central role of training data in shaping model capabilities and motivate the need for mid-training (\S~\ref{subsec:midtraining}) as a practical means of introducing such data, enabling subsequent RL training to operate over richer priors.


In this work, we investigate how mid-training with diverse self-generated data can be a simple and effective way to improve RL in language models. Specifically, we study a simple setting in which, for each question, we generate multiple diverse, correct solutions and use them to perform fine-tuning as a mid-training step. In our setup, training data is generated directly from the same base model that later undergoes this mid-training stage, enabling self-improvement from its own generated reasoning without relying on distillation from stronger models. 
We adopt a bootstrapped synthetic-data generation framework that systematically produces diverse yet correct reasoning trajectories by conditioning on explicit problem-solving heuristics. Inspired by the classical problem-solving approaches of George Pólya \citep{polyahowtosolveit}, we construct heuristic-guided few-shot prompts that explicitly encourage the model to generate responses with distinct reasoning approaches for the same underlying question. By filtering these responses for correctness and adherence to the targeted heuristic, we obtain multiple diverse solution variants per question without requiring human annotation or supervision from stronger models. Using these self-generated reasoning trajectories, we fine-tune the base model. 


Our experiments show that applying GRPO-based RL to mid-trained models consistently improves performance across a range of mathematical reasoning benchmarks, with gains often increasing as the diversity of reasoning approaches introduced during mid-training grows. Compared to vanilla RL baselines, models mid-trained with a larger number of heuristics achieve higher pass@\(\!k\), particularly at larger values of \(k\). At pass@\(\!64\), this corresponds to improvements of 2.85\% on MATH-500, 5.7\% on AIME 2024, 6.55\% on AIME 2025, 2.26\% on HMMT 2025, 6.34\% on AMC 2023, and 3.17\% on OlympiadBench. While the optimal number of heuristics varies across datasets, models trained with higher reasoning diversity consistently outperform both vanilla RL and STaR across all benchmarks. 

Additionally, we provide a theoretical perspective that explains how mid-training with multiple approaches helps RL and explain how it incentivizes combining reasoning strategies in a single response. This compositional behavior is empirically observed in our analysis in \S~\ref{subsec:analysisofreasoningtraces} and aligns with findings from recent research \citep{yuan2025fxgxfgxllms, cheng2025atomic}. Our additional analysis shows that, under a fixed mid-training budget, learning multiple problem-solving approaches per question yields greater benefits than learning a single approach across a larger number of questions. We further show that our heuristic-guided self-generated data is more diverse than data distilled from a stronger teacher model (Vendi Score 13.81 vs. 10.95 for QwQ-32B distillation), leading to improved downstream pass@1 and pass@64 performance. Finally, despite Pólya’s heuristics having a math-centric focus, we observe that the resulting problem-solving strategies generalize beyond math, yielding considerable gains on out-of-domain benchmarks, including around code generation and narrative reasoning.

\section{Related Works}

\paragraph{Synthetic Data Generation} A large body of work uses LLMs to generate synthetic data without human annotation~\citep{unnatural_inst, targen,wizardlm}. Earlier work focused on generating training datasets for instruction tuning~\citep{crosstask_gen,flan_instruction}; \citet{Self-Instruct} proposes a bootstrapping framework in which a pre-trained model iteratively generates instruction-response pairs to improve its instruction-following ability. Other works~\citep{Orca, orca2} improve the reasoning capabilities of smaller models by distilling synthetic data from larger models. \citet{STaR} introduces a self-training approach in which a model fine-tunes on its own generated rationales, and \citet{RL-Incorrect} shows that using both correct and incorrect synthetic traces improves RL sample efficiency on math reasoning. \citet{how_many_prompts} show that increasing instruction variation yields larger gains than scaling the number of training instances. Our work differs from these prior approaches by generating multiple response variations per question via heuristic-conditioned prompting and studying the effect of this diversity during mid-training on subsequent RL.

\paragraph{Reinforcement Learning for LLMs} Reinforcement Learning from Human Feedback (RLHF)~\citep{rlhf} has become a standard post-training step, aligning models with human preferences by training a reward model and optimizing the policy with algorithms such as Proximal Policy Optimization (PPO)~\citep{ppo}. Direct Preference Optimization (DPO)~\citep{dpo} reframes preference alignment as a supervised objective and matches or exceeds PPO-based methods, with several variants further modifying the training objective~\citep{step_dpo,TPO,offsetdpo,SIMPO}. Reinforcement Learning with Verifiable Rewards (RLVR)~\citep{rlvr_lambert,deepmath,deepseekr1} has since driven substantial gains on challenging reasoning tasks. However, RL is more effective when applied to base models with strong priors~\citep{think_tuning,cognitive_behaviors,zhang2025interplaypretrainingmidtrainingrl}, and several works~\citep{olmo2025olmo} distill task-specific reasoning data before RL. \citet{yuan2025fxgxfgxllms} further show that RL enables models to learn novel compositions of atomic skills not explicitly taught during prior training. Our work differs by mid-training on diverse self-generated responses produced via Pólya-style problem-solving heuristics, and theoretically analyzing how subsequent RL combines these approaches.

\section{Preliminaries}




\paragraph{Supervised Fine-Tuning (SFT).}
SFT trains a pre-trained LLM \(\pi_\theta\) on a dataset of questions and their corresponding responses. Let
\(\mathcal{D}_{\mathrm{SFT}} = \{(x, y)\}\) denote the training dataset, where \(x\) is a question and \(y = (y_1, y_2, \ldots,
y_T)\) is the corresponding response sequence. SFT minimizes the negative log-likelihood of the responses in
\(\mathcal{D}_{\mathrm{SFT}}\),
{\small
\[
\mathcal{L}_{\mathrm{SFT}}(\theta)
= - \mathbb{E}_{(x,y)\sim \mathcal{D}_{\mathrm{SFT}}}
\left[ \sum_{t=1}^{T} \log \pi_\theta \bigl( y_t \mid y_{<t}, x \bigr) \right].
\]
}

\paragraph{RL via Policy Gradients.}
In RL, the language model \(\pi_\theta\) serves as a policy that defines a conditional distribution over output tokens given a
question \(x\) and the previously generated tokens. Let \(\mathcal{D}_{\mathrm{RL}}\) denote a dataset of questions used for RL.
For a given question \(x \sim \mathcal{D}_{\mathrm{RL}}\), the model generates a response \(y = (y_1, \ldots, y_T)\)
autoregressively by sampling tokens according to \(\pi_\theta(y_t \mid y_{<t}, x)\). At each time step \(t\), the state is \((x,
y_{<t})\) and the action is the next token \(y_t\), so a complete response constitutes a trajectory sampled from the policy. The
objective is to maximize the expected reward of trajectories generated for questions in \(\mathcal{D}_{\mathrm{RL}}\), with
gradient
{\small
\[
\nabla_\theta J(\theta)
=
\mathbb{E}_{x \sim \mathcal{D}_{\mathrm{RL}},\, y \sim \pi_\theta(\cdot \mid x)}
\left[
A(x, y)\, \nabla_\theta \log \pi_\theta(y \mid x)
\right],
\]
}

where \(A(x, y)\) is an advantage function that measures the relative quality of a response under the reward signal. Among
commonly used approaches such as REINFORCE~\citep{REINFORCE} and PPO~\citep{ppo}, we employ Group Relative Policy Optimization
(GRPO)~\citep{grpo} in our experiments due to its simplicity.


\section{Mid-Training (MT)} \label{subsec:midtraining} Prior works~\citep{mo2025mid,tu2025survey} characterize mid-training as an intermediate phase between pre-training and post-training, in which a base model is further trained on high-quality or task-focused data to induce targeted improvements while preserving existing capabilities. We focus on mid-training with diverse self-generated data and study its role in improving subsequent RL training. Specifically, we construct a mid-training dataset in which each question $x$ is associated with multiple correct solution trajectories, exposing the model to $n$ valid ways of solving the same problem during supervised training. This setup lets us analyze how the \textit{guided} diversity introduced during mid-training influences both RL performance and the reasoning approaches observed during the RL phase.

\paragraph{Heuristic-Guided Data Construction}
\begin{figure*}[t]
\begin{center}
\includegraphics[width=0.9\linewidth]{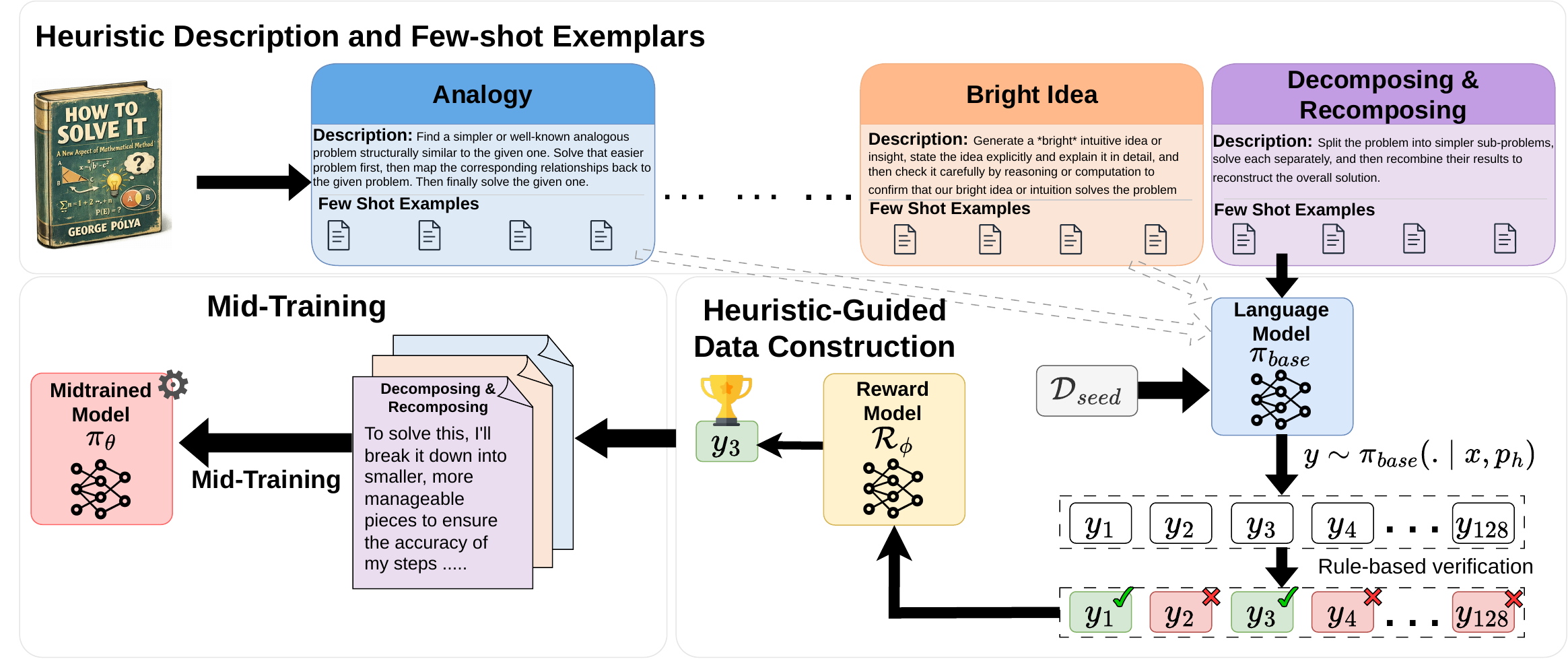}
\end{center}
\caption{\textbf{Overview of heuristic-guided mid-training.} 
\textit{Top:} We draw on Pólya's problem-solving heuristics, each defined by a textual description and few-shot exemplars demonstrating the heuristic applied to example problems. 
\textit{Bottom right:} For each question--heuristic pair, we sample 128 candidate responses from the base model conditioned on the heuristic prompt, filter for correctness via rule-based verification, and select the highest-scoring response according to a reward model $\mathcal{R}_\phi$. 
\textit{Bottom left:} The resulting dataset $\mathcal{D}_{\text{Pólya}}$, containing $n$ solution variants per question, is used to mid-train the base-instruct model before RL.}
\label{fig:datagen}
\end{figure*}

To systematically introduce diverse ways of solving the same problem, we draw inspiration from the problem-solving heuristics
described in \textit{How to Solve It} by George Pólya~\citep{polyahowtosolveit}. This work presents a structured view of problem
solving, emphasizing that a single problem can often be approached through multiple valid strategies, such as working backward,
introducing auxiliary quantities, or decomposing the problem into simpler subproblems. A comprehensive list of heuristic
descriptions and examples is provided in \S~\ref{appendix:heuristics}.

Formally, we associate each heuristic \(h\) with a textual description \(D_h\) and a small set of few-shot exemplars
\(\mathcal{E}_h = \{(x_i, r_{h,i}, a_i)\}_{i=1}^{4}\). Given an existing dataset of questions and answers
\(\mathcal{D}_{\mathrm{seed}} = \{(x_i, a_i)\}_{i=1}^{|\mathcal{D}_{\mathrm{seed}}|}\), our goal is to generate, for each question
\(x\), a set of solution trajectories that reflect distinct reasoning behaviors. To this end, for each question--heuristic pair
\((x, h)\), we prompt the base model \(\pi\) with \(D_h\) and \(\mathcal{E}_h\) and sample multiple candidate responses.

The sampled trajectories are then filtered and refined to ensure both correctness and adherence to the targeted heuristic. We
first apply rule-based verifiers (e.g., \texttt{Math-Verify}) to retain responses that produce the correct final answer. The
remaining candidates are scored using a reward model \(\mathcal{R}_\phi\), which evaluates how well each response follows the
intended heuristic given \(x\) and \(D_h\). For each question--heuristic pair, we select the response with the highest reward
score as the final heuristic-specific solution trajectory. The resulting collection constitutes the mid-training data. Training on
this dataset differs from standard SFT in that each question \(x\) is paired with multiple distinct solution trajectories rather
than a single correct response. Let \(\mathcal{Y}_x = \{y^{(1)}, \ldots, y^{(n)}\}\) denote the set of heuristic-specific
trajectories associated with question \(x\). The mid-training objective is the average negative log-likelihood over these
trajectories,
{\small
\[
\mathcal{L}_{\mathrm{MT}}(\theta)
= - \mathbb{E}_{(x,\mathcal{Y}_x)}
\left[
\frac{1}{n} \sum_{i=1}^{n} \sum_{t=1}^{T_i}
\log \pi_\theta\bigl(y^{(i)}_t \mid y^{(i)}_{<t}, x\bigr)
\right].
\]
}
Optimizing this objective encourages the model to assign non-negligible probability to several next tokens at prefixes shared
across different solution strategies, each corresponding to a distinct reasoning behavior. As a result, mid-trained models can
exhibit next-token distributions with multiple high-probability modes at such branching points, in contrast to models trained on a
single solution per question. In the following section, we analyze how these $n$-modal next-token distributions affect
policy-gradient updates during RL.

\subsection{Theory}
\label{subsec:theory}

\paragraph{Theorem 4.1} Let $\pi_{\theta}(.\;|\; x,y_{<t}) = \operatorname{softmax}(z_{\pi}(x,y_{<t}))$ be an auto-regressive language model parameterized by $\theta$ updated via a single policy gradient step with learning rate $\eta$ and advantage $A$ for a sampled token $y_t\in\mathcal{V}$. Let $\epsilon$ denote the residual probability mass assigned to the tail of the distribution. Under a uni-modal regime where $\pi_{\theta}(y_{t}\;|\;x,y_{<t}) = 1-\epsilon$; and an $N$-modal regime where the mass is split among N dominant modes ($\pi_{\theta}(y_{t}\;|\;x,y_{<t}) \approx \frac{1-\epsilon}{N}$), the expected first order change in the probability of the sampled token scales as follows:
{\small
\[
\Delta \pi_{\theta}(y_{t}\mid x,y_{<t})
\approx
\begin{cases}
\eta A\epsilon^2, & \text{uni-modal},\\
\eta A\frac{1}{N}\!\left(1-\frac{1}{N}\right)
, & \text{$N$-modal}.
\end{cases}
\]
}

\paragraph{Key insight.} The above theorem highlights how a single policy-gradient update affects the probability distribution of a language model. We consider two regimes: \textit{uni-modal}: when a model is too confident on a single token and \textit{N-modal}: when a model has high probability on N tokens. For the purposes of this discussion, we assume that in the $N$-modal regime probability mass is distributed across $N << |\mathcal{V}|$ dominant tokens, each with probability $\approx\frac{1-\epsilon}{N}$, although in practice these probabilities need not be uniform. Implicitly, our mid-training approach incentivizes such N-modal next-token distributions as discussed in the previous section. From the theorem, we see that under a uni-modal regime, the induced change in the sampled token's probability scales as $\epsilon^2$. Consequently, once the model becomes highly confident in its prediction, a single policy-gradient update produces only a negligible change in the output distribution. In contrast, under the N-modal regime, the sampled token's probability scales as $\frac{1}{N}$, which leads to conservative yet appreciable changes. As N increases, the changes become more conservative, preventing it from committing to a single mode within few update steps. This prevents mode collapse, as discussed in recent work~\citep{yue2025reinforcement}. 



\paragraph{Proposition 4.2}
Consider the $N$-modal regime in which probability mass $1-\epsilon$ is distributed among $N \ll |\mathcal{V}|$ dominant modes. Under a single policy-gradient update with negative advantage $A<0$, the probability mass removed from the sampled token $y_t$ is redistributed predominantly to the remaining $N-1$ dominant modes.

\paragraph{Discussion} Proposition 4.2 highlights how probability mass is redistributed during a single negative policy gradient update. This result provides a conceptual explanation for why exposing the model to diverse approaches to solving a problem can be beneficial. In particular, when multiple high-probability continuations exist at a given prefix, the sampled token corresponds to one specific continuation, while other tokens correspond to alternative plausible continuations (i.e., different problem-solving approaches). A negative update reduces the probability of the sampled token and reallocates mass toward these alternatives. As a result, the model learns to combine different problem-solving approaches into a single response during RL training, as shown in \S~\ref{subsec:qualexample}. As the model is exposed to more problem-solving approaches, it increasingly learns to combine elements of these approaches within a single response, through the course of RL training.

\section{Experiments}
\label{sec:experiments}

\subsection{Experimental Setup}
\paragraph{Baselines}
In our investigation, we compare our models against a zero-shot baseline and STaR \citep{STaR}. The zero-shot baseline evaluates the base model without any additional training. For experiments involving RL, we compare mid-trained models against two baselines. The first is a vanilla RL baseline in which GRPO is applied directly to the base model. The second is STaR+RL, where GRPO is applied after STaR-based fine-tuning.


\paragraph{Training Datasets}
For mid-training, we use the GSM8K \citep{gsm8k} training set, which contains 7{,}473 questions--answer pairs and serves as the seed dataset $\mathcal{D}_{\text{seed}}$. Following the procedure described in \textsection \ref{subsec:midtraining}, we obtain a filtered set of 7{,}112 questions, each paired with up to 64 heuristic-specific variants of the correct solution. We construct multiple mid-training datasets with $n \in \{1, 2, 4, \ldots, 64\}$ solution variants per question and train a separate model on each mid-training dataset.

\paragraph{Models and Reward}
We use \textbf{Llama~3.2--3B--Instruct} \citep{Llama3-herd} as the primary base model for all experiments, including both baselines and mid-training, and report additional results with \textbf{Qwen2.5--7B--Instruct} in the \S~\ref{app:qwen_results}. We use \textbf{Skywork-Reward-V2-Llama-3.2-3B} \citep{reward_model_skywork} as the reward model ($\mathcal{R}_{\phi}$) to score responses during data generation.

\paragraph{Evaluation Details}
We evaluate on six mathematical reasoning benchmarks: \textbf{Math-500}~\citep{math-500}, \textbf{AIME 2024}~\citep{aime24}, \textbf{AIME 2025}~\citep{aime25}, \textbf{AMC 2023}~\citep{amc2023}, \textbf{HMMT 2025}~\citep{hmmt-25-matharena}, and \textbf{OlympiadBench}~\citep{olympiadbench}, covering a wide range of difficulties and reasoning types. We use \texttt{Math-Verify}~\citep{huggingface_math_verify_2025} to verify the correctness of the models' generated solutions automatically. 
Model performance is measured using the pass@$k$ metric \citep{passatcode}, where a problem is solved if at least one of $k$ samples is correct. Following~\citet{humaneval}, we use the unbiased estimator $\mathrm{pass@}k = \mathbb{E}\bigl[1 - \binom{n-c}{k} / \binom{n}{k}\bigr]$, where $n$ is the total number of samples and $c$ the number correct.


\subsection{Results Discussion}
\renewcommand{\arraystretch}{1.18}
\setlength{\extrarowheight}{0.8pt}
\newcommand{\se}[1]{{\tiny\raisebox{0.4ex}{$\pm$#1}}}

\begin{table*}[htbp!]
\caption{
\textbf{Mid-Training Results.}
Pass@1 and pass@64 performance on six mathematical reasoning benchmarks for Llama~3.2--3B--Instruct.
$\mathcal{D}_{\text{Pólya}}$ denotes mid-training on GSM8K using \emph{self-generated}, diverse, heuristic-guided solution variants, with $n$ variants per question.
Results are reported as mean $\pm$ standard error (SE) over 3 (inference) runs.
Best and second-best results are \textbf{bolded} and \underline{underlined}.
}
\label{tab:sft_results}

\centering
\resizebox{\textwidth}{!}{%
\begin{tabular}{@{}>{\centering\arraybackslash}m{1.8cm}|>{\centering\arraybackslash}m{0.7cm}|cccccc!{\vrule}c@{}}
\toprule
\multirow{2}{*}{\textbf{Methods}} &
\multirow{2}{*}{\textit{n}} &
\multicolumn{6}{c!{\vrule}}{\textit{Reasoning Benchmarks} \scriptsize{(pass@1 / pass@64)}} &
\multirow{2}{*}{\textsc{Avg.}} \\
\cmidrule(lr){3-8}
& &
\textsc{Math-500} &
\textsc{AIME 24} &
\textsc{AIME 25} &
\textsc{AMC 23} &
\textsc{HMMT 25} &
\textsc{OlympiadBench} & \\
\midrule
Zero-Shot & N/A &
$37.34\se{0.03}$ / $87.76\se{0.21}$ &
$2.81\se{0.04}$ / $32.08\se{0.14}$ &
$0.26\se{0.02}$ / $12.84\se{0.83}$ &
$16.95\se{0.05}$ / $83.49\se{0.67}$ &
$1.26\se{0.00}$ / $19.48\se{0.14}$ &
$7.86\se{0.28}$ / $42.13\se{0.30}$ &
$11.08$ / $46.30$ \\
\addlinespace[2pt]
$\mathcal{D}_{\text{STaR}}$ & N/A &
$\best{41.74}\se{0.23}$ / $86.91\se{0.18}$ &
$\best{4.39}\se{0.04}$ / $30.44\se{0.16}$ &
$0.33\se{0.01}$ / $13.91\se{0.32}$ &
$\best{20.50}\se{0.03}$ / $\best{86.75}\se{0.10}$ &
$1.37\se{0.01}$ / $18.08\se{0.11}$ &
$\best{9.78}\se{0.04}$ / $41.83\se{0.15}$ &
$\best{13.02}$ / $46.32$ \\
\midrule
\multirow{7}{*}{$\mathcal{D}_{\text{Pólya}}$}
 & 1 &
$31.03\se{0.07}$ / $88.11\se{0.10}$ &
$1.61\se{0.04}$ / $30.72\se{0.41}$ &
$0.38\se{0.01}$ / $16.71\se{0.81}$ &
$14.24\se{0.10}$ / $82.79\se{0.11}$ &
$1.44\se{0.02}$ / $\best{20.86}\se{0.16}$ &
$6.80\se{0.01}$ / $42.35\se{0.22}$ &
$9.25$ / $46.92$ \\
\addlinespace[2pt]
 & 2 &
$34.01\se{0.04}$ / $88.08\se{0.05}$ &
$1.82\se{0.02}$ / $31.42\se{0.41}$ &
$\second{0.40}\se{0.01}$ / $17.44\se{0.65}$ &
$16.08\se{0.06}$ / $84.45\se{0.17}$ &
$1.51\se{0.02}$ / $20.34\se{0.37}$ &
$7.34\se{0.01}$ / $42.57\se{0.22}$ &
$10.19$ / $47.38$ \\
\addlinespace[2pt]
 & 4 &
$34.88\se{0.04}$ / $88.37\se{0.13}$ &
$1.83\se{0.04}$ / $31.04\se{0.33}$ &
$0.37\se{0.01}$ / $16.70\se{0.50}$ &
$16.36\se{0.01}$ / $83.51\se{0.32}$ &
$1.50\se{0.02}$ / $19.77\se{0.48}$ &
$7.48\se{0.05}$ / $42.82\se{0.15}$ &
$10.40$ / $47.04$ \\
\addlinespace[2pt]
 & 8 &
$36.17\se{0.07}$ / $88.53\se{0.13}$ &
$1.99\se{0.04}$ / $31.20\se{0.04}$ &
$0.34\se{0.05}$ / $15.26\se{1.52}$ &
$16.79\se{0.14}$ / $83.19\se{0.61}$ &
$1.54\se{0.04}$ / $20.16\se{0.32}$ &
$7.72\se{0.01}$ / $42.47\se{0.17}$ &
$10.76$ / $46.80$ \\
\addlinespace[2pt]
 & 16 &
$37.69\se{0.05}$ / $\second{88.75}\se{0.27}$ &
$2.35\se{0.02}$ / $32.06\se{0.58}$ &
$\second{0.40}\se{0.01}$ / $\second{17.52}\se{0.16}$ &
$17.55\se{0.01}$ / $84.03\se{0.65}$ &
$1.58\se{0.04}$ / $\second{20.77}\se{0.33}$ &
$8.15\se{0.01}$ / $42.87\se{0.15}$ &
$11.29$ / $47.67$ \\
\addlinespace[2pt]
 & 32 &
$37.94\se{0.05}$ / $88.53\se{0.07}$ &
$\second{2.64}\se{0.04}$ / $\best{33.27}\se{0.46}$ &
$\second{0.40}\se{0.02}$ / $16.97\se{0.67}$ &
$\second{17.84}\se{0.14}$ / $\second{85.18}\se{0.37}$ &
$\second{1.63}\se{0.02}$ / $20.37\se{0.40}$ &
$8.45\se{0.06}$ / $\best{43.57}\se{0.17}$ &
$11.48$ / $\second{47.98}$ \\
\addlinespace[2pt]
 & 64 &
$\second{38.11}\se{0.08}$ / $\best{88.94}\se{0.10}$ &
$2.48\se{0.05}$ / $\second{32.32}\se{0.48}$ &
$\best{0.44}\se{0.02}$ / $\best{18.66}\se{0.69}$ &
$17.64\se{0.06}$ / $85.01\se{0.42}$ &
$\best{1.69}\se{0.06}$ / $20.62\se{0.51}$ &
$\second{8.66}\se{0.03}$ / $\second{43.46}\se{0.08}$ &
$\second{11.50}$ / $\best{48.17}$ \\
\bottomrule
\end{tabular}%
}
\end{table*}

\input{figures/pass_at_k_tiks}
\paragraph{Results of Mid-Training}

From Table~\ref{tab:sft_results}, we observe that mid-training with self-generated data guided by Pólya’s heuristics consistently improves performance across benchmarks. At pass@\(\!1\), mid-trained models show steady gains as the number of heuristics increases, with the average pass@\(\!1\) improving from 9.25\% at \(n=1\) to 11.50\% at \(n=64\). While these gains are modest, STaR attains a higher average pass@\(\!1\) of 13.02\%, corresponding to a +1.94\% improvement over the zero-shot baseline. In contrast, under pass@\(\!64\), STaR provides little improvement relative to zero-shot performance. We hypothesize that this may stem from the presence of a narrow set of similar reasoning patterns that are frequently correct during data construction, which encourages the model to focus on a limited set of problem-solving approaches. This observation also motivates our decision to use only a single STaR iteration in our experiments to mitigate potential overfitting.

In contrast, the benefits of our specific mid-training are more pronounced under pass@\(\!64\). On Math-500, pass@\(\!64\) improves from 87.76\% in the zero-shot setting to 88.94\% at \(n=64\). Larger gains are observed on more challenging benchmarks, including AIME~2025 (12.84\% \(\rightarrow\) 18.66\%), AMC~2023 (83.49\% \(\rightarrow\) 85.18\%), and OlympiadBench (42.13\% \(\rightarrow\) 43.57\%). These trends are reflected in the average pass@\(\!64\), which improves from 46.30\% for the zero-shot baseline and 46.32\% for STaR to 48.17\% at \(n=64\). Taken together, the results indicate that mid-training with diverse, self-generated reasoning encourages the model to learn a broader set of problem-solving strategies.

\paragraph{Results with RL} Fig.~\ref{fig:pass_at_k} shows the effect of GRPO-based RL on models initialized with different numbers of Pólya-style approaches. At pass@$2$, mid-trained models generally outperform the vanilla RL baseline. On MATH-500, vanilla RL and STaR+RL reach 56.48\% and 58.13\% respectively, while mid-trained models reach 59--60\% for several values of $n$ (e.g., 59.68\% at $n=32$).  Although performance at pass@$2$ varies with the number of heuristics and is not strictly monotonic, models initialized with multiple heuristics consistently achieve higher pass@$2$ values than the vanilla RL.

The gap widens as $k$ grows, with STaR+RL typically lying between vanilla RL and mid-trained models. At pass@$64$ on MATH-500, vanilla RL reaches 83.42\% versus 86.27\% for mid-training with $n=64$. Larger gains appear on harder benchmarks, e.g., AIME~2025 (16.91\% $\rightarrow$ 23.34\%) and AMC~2023 (78.18\% $\rightarrow$ 84.52\%). Averaged across benchmarks, vanilla RL reaches 44.21\% and STaR+RL 45.69\% at pass@$64$, while mid-trained models reach 48.09\% ($n=16$) and 47.62\% ($n=64$). In some benchmarks, $n=16$ outperforms $n=64$. We hypothesize this is tied to the rollout group size $g=16$ matching $n$: when $g=n$, based on rollout logs earlier in the training, each group covers all or most of the $n$ learned strategies, whereas when $g<n$, groups capture only subsets, limiting RL's ability to leverage learned diverse approaches. Our Qwen2.5--7B--Instruct experiments (\S~\ref{app:qwen_results}, $g=8$) show the same pattern, with $n=8$ best or on par with larger $n$. Overall, our study shows that mid-training
with self-generated data improves subsequent RL training.

\section{Analysis and Discussion}
\label{sec:analysis}

\subsection{Analysis of Reasoning Traces}
\label{subsec:analysisofreasoningtraces}

The results discussed in the previous section indicate that mid-training on self-generated data with diverse problem-solving approaches improves subsequent
RL. Our theoretical discussion in \textsection\ref{subsec:theory} offers insight into how these models might combine different
ways of solving a problem within a single reasoning chain during RL.
\begin{figure}[t]
    \centering

    \begin{subfigure}[t]{0.48\textwidth}
        \centering
        \resizebox{\linewidth}{!}{\begin{tikzpicture}
    \begin{axis}[
        width=\textwidth,
        height=3.7cm,
        xlabel={$n$ (trained behaviors)},
        ylabel={Comp. rate (\%)},
        xmin=0, xmax=18,
        ymin=0, ymax=70,
        xtick={2, 4, 8, 16},
        ytick={0, 20, 40, 60},
        legend style={
            at={(0.5,1.02)},
            anchor=south,
            legend columns=2,
            draw=none,
            fill=none,
            font=\small,
            column sep=8pt,
        },
        grid=major,
        grid style={gray!30},
        mark size=3pt,
    ]
    
    \addplot[
        color=blue!50,
        mark=square*,
        thick,
        mark options={fill=blue!30},
    ] coordinates {
        (2, 10.0)
        (4, 23.3)
        (8, 23.3)
        (16, 23.3)
    };
    
    \addplot[
        color=blue!90,
        mark=*,
        thick,
        mark options={fill=blue!70},
    ] coordinates {
        (2, 0.0)
        (4, 10.0)
        (8, 16.7)
        (16, 56.7)
    };
    
    \legend{Mid-Trained, Mid-Trained+RL}
    
    \draw[<->, gray, thick] (axis cs:16,23.3) -- (axis cs:16,56.7)
        node[midway, right, font=\scriptsize] {+34\%};
    
    \end{axis}
\end{tikzpicture}}
        \caption{Composition rate across mid-training scales.}
        \label{fig:composition_rate}
    \end{subfigure}
    \hfill
    \begin{subfigure}[t]{0.48\textwidth}
        \centering
        \resizebox{\linewidth}{!}{\begin{tikzpicture}[
      behavior/.style={font=\small\bfseries, align=right, text=black!90},
      combo/.style={font=\small\bfseries, align=left},
      transition/.style={font=\small\bfseries, text=gray!90},
      xscale=1,
      yscale=0.9,
      scale=0.88, transform shape
  ]

  \node[behavior] (auxelem) at (0, 5.4) {Aux.Elem};
  \node[behavior] (condition) at (0, 4.5) {Condition};
  \node[behavior] (restate) at (0, 3.6) {Restate};
  \node[behavior] (bolzano) at (0, 2.7) {Bolzano};
  \node[behavior] (decompose) at (0, 1.8) {Decompose};
  \node[behavior] (carryout) at (0, 0.9) {CarryOut};

  \node[combo] (bd) at (5.5, 5.8) {Bolz. $+$ Decomp.};
  \node[transition] at (5.5, 5.45) {0\% $\rightarrow$ \textbf{37\%}};

  \node[combo] (rd) at (5.5, 4.9) {Rest. $+$ Decomp.};
  \node[transition] at (5.5, 4.55) {0\% $\rightarrow$ \textbf{30\%}};

  \node[combo] (brd) at (5.5, 3.8) {Bolz. $+$ Rest. $+$ Decomp.};
  \node[transition] at (5.5, 3.45) {0\% $\rightarrow$ \textbf{23\%}};

  \node[combo] (bcd) at (5.5, 2.9) {Bolz. $+$ Carry. $+$ Decomp.};
  \node[transition] at (5.5, 2.55) {0\% $\rightarrow$ \textbf{20\%}};

  \node[combo] (bcrd) at (5.5, 1.8) {Bolz. $+$ Carry. $+$ Rest. $+$ Decomp.};
  \node[transition] at (5.5, 1.45) {0\% $\rightarrow$ \textbf{17\%}};

  \node[combo] (abcd) at (5.5, 0.9) {Aux. $+$ Bolz. $+$ Cond. $+$ Decomp.};
  \node[transition] at (5.5, 0.55) {0\% $\rightarrow$ \textbf{13\%}};

  \draw[gray!30, thin] (4.2, 4.35) -- (7.8, 4.35);
  \draw[gray!30, thin] (4.2, 2.35) -- (7.8, 2.35);

  \definecolor{c1}{HTML}{2E86AB}
  \definecolor{c2}{HTML}{4ECDC4}
  \definecolor{c3}{HTML}{F7B801}
  \definecolor{c4}{HTML}{F18701}
  \definecolor{c5}{HTML}{D64550}
  \definecolor{c6}{HTML}{7B2D8E}

  \draw[c6, line width=1.3pt, opacity=0.7] (auxelem.east) to[out=0, in=170] (abcd.west);
  \draw[c6, line width=1.3pt, opacity=0.7] (condition.east) to[out=0, in=165] (abcd.west);
  \draw[c6, line width=1.3pt, opacity=0.7] (bolzano.east) to[out=-5, in=175] (abcd.west);
  \draw[c6, line width=1.3pt, opacity=0.7] (decompose.east) to[out=-10, in=185] (abcd.west);

  \draw[c5, line width=1.7pt, opacity=0.7] (restate.east) to[out=-5, in=160] (bcrd.west);
  \draw[c5, line width=1.7pt, opacity=0.7] (bolzano.east) to[out=-3, in=170] (bcrd.west);
  \draw[c5, line width=1.7pt, opacity=0.7] (decompose.east) to[out=5, in=190] (bcrd.west);
  \draw[c5, line width=1.7pt, opacity=0.7] (carryout.east) to[out=10, in=200] (bcrd.west);

  \draw[c4, line width=2.0pt, opacity=0.7] (bolzano.east) to[out=0, in=175] (bcd.west);
  \draw[c4, line width=2.0pt, opacity=0.7] (decompose.east) to[out=10, in=195] (bcd.west);
  \draw[c4, line width=2.0pt, opacity=0.7] (carryout.east) to[out=20, in=210] (bcd.west);

  \draw[c3, line width=2.3pt, opacity=0.7] (restate.east) to[out=5, in=185] (brd.west);
  \draw[c3, line width=2.3pt, opacity=0.7] (bolzano.east) to[out=8, in=190] (brd.west);
  \draw[c3, line width=2.3pt, opacity=0.7] (decompose.east) to[out=15, in=200] (brd.west);

  \draw[c2, line width=3.0pt, opacity=0.7] (restate.east) to[out=12, in=195] (rd.west);
  \draw[c2, line width=3.0pt, opacity=0.7] (decompose.east) to[out=25, in=210] (rd.west);

  \draw[c1, line width=3.7pt, opacity=0.75] (bolzano.east) to[out=15, in=200] (bd.west);
  \draw[c1, line width=3.7pt, opacity=0.75] (decompose.east) to[out=30, in=215] (bd.west);

  \foreach \node in {auxelem, condition, restate, bolzano, decompose, carryout}
      \fill[black!60] (\node.east) circle (1.5pt);

  \end{tikzpicture}}
        \caption{Emergence of combinations at $n=16$.}
        \label{fig:sankey_combinations}
    \end{subfigure}

    \caption{Behavior composition after mid-training and reinforcement learning. Left: composition rate as the number of trained behaviors increases. Right: the most common multi-behavior combinations emerging after RL at $n=16$.}
    \label{fig:composition_combined}
\end{figure}
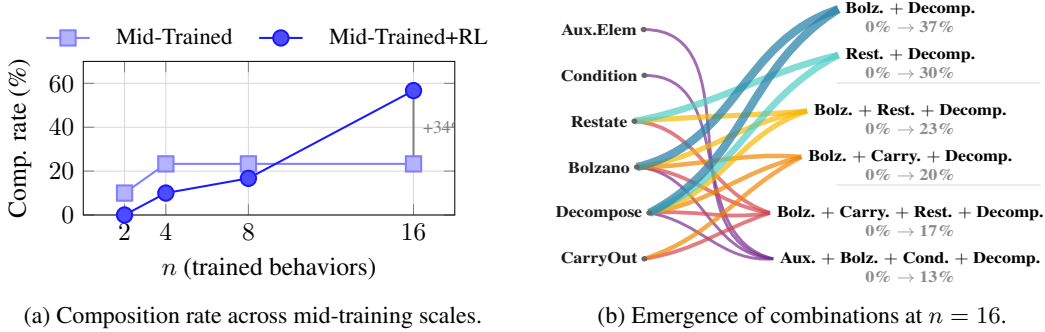

To study this effect, we analyze reasoning traces sampled from both RL and mid-trained models using an LLM-based judge (GPT-4o-mini; \citep{hurst2024gpt}) to identify
problem-solving approaches in the generated solutions. The classifier is defined using few-shot prompts derived from 64 of the
reasoning heuristics in Pólya's taxonomy, each consisting of a heuristic description along with positive and negative examples
(see Fig.~\ref{fig:heuristic_prompt}). Given a reasoning chain, the judge outputs a binary decision for each heuristic. We apply
this to chains sampled from mid-trained models trained with \(n \in \{2, 4, 8, 16\}\) distinct heuristics per question and their
RL-trained counterparts, randomly selecting one solution per problem from AIME 2024. Fig.~\ref{fig:composition_rate} shows the
fraction of reasoning chains that exhibit more than one problem-solving behavior. For mid-trained models without RL, this fraction
remains relatively stable, whereas RL leads to more chains that combine multiple approaches, and the trend strengthens as the
number of approaches introduced during mid-training increases. When $n=16$, the RL-trained model exhibits composition in 56.7\% of
sampled chains, compared to 23.3\% for the corresponding mid-trained model.

To better understand how these approaches are combined, Figure~\ref{fig:sankey_combinations} lists the most common heuristic
combinations observed for models trained with $n=16$. Such combinations are not explicitly demonstrated during mid-training, but
emerge consistently after RL. This provides empirical evidence that RL combines problem-solving approaches introduced during
mid-training, in line with our theoretical discussion and recent work~\citep{yuan2025fxgxfgxllms, cheng2025atomic}. These findings
raise an interesting question. \textit{Does RL truly give rise to emergent behaviors in large language models, or does it
primarily compose existing behaviors learned during pre-training, which may appear emergent due to limited visibility into the
pre-training data?} We leave a more detailed investigation to future work.

\definecolor{inkblue}{RGB}{56,108,176}
\definecolor{teal}{RGB}{77,190,182}
\definecolor{gold}{RGB}{241,196,15}
\definecolor{mutedred}{RGB}{214,69,65}
\definecolor{vanillared}{RGB}{255,150,150}

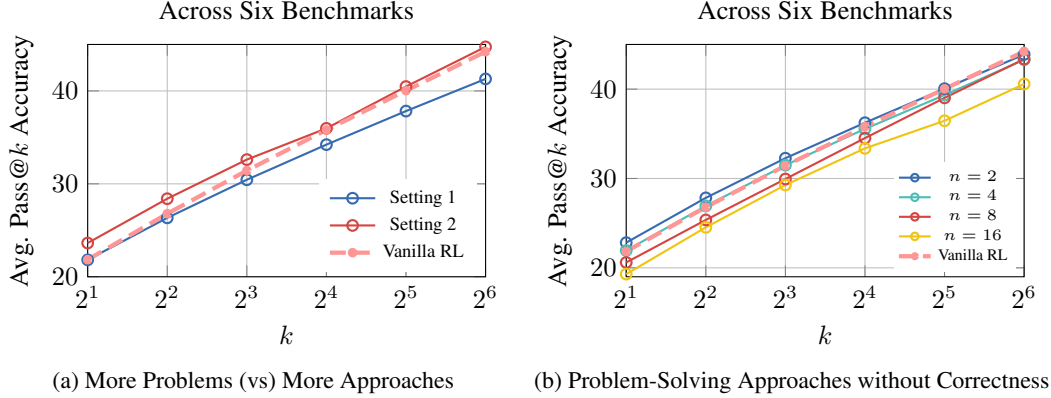
\begin{figure}[t]
\centering
\begin{subfigure}[t]{0.49\textwidth}
\centering
\begin{tikzpicture}
\begin{axis}[
    title={Across Six Benchmarks},
    xlabel={$k$},
    ylabel={Avg. Pass@$k$ Accuracy},
    xmode=log,
    log basis x=2,
    xmin=2, xmax=64,
    xtick={2,4,8,16,32,64},
    ymin=20, ymax=45,
    ymajorgrids=true,
    xmajorgrids=true,
    width=\linewidth,
    height=0.68\linewidth,
    legend style={
        draw=none,
        font=\scriptsize,
        legend columns=1,
        at={(0.98,0.02)},
        anchor=south east
    }
]
\addplot[color=inkblue, thick, mark=o]
coordinates {(2,21.82)(4,26.34)(8,30.44)(16,34.22)(32,37.83)(64,41.29)};
\addlegendentry{Setting 1}
\addplot[color=mutedred, thick, mark=o]
coordinates {(2,23.63)(4,28.40)(8,32.61)(16,35.97)(32,40.47)(64,44.74)};
\addlegendentry{Setting 2}
\addplot[
    color=vanillared,
    ultra thick,
    dash pattern=on 6pt off 3pt,
    mark=*,
    mark size=1pt
] coordinates {(2,21.84)(4,26.79)(8,31.43)(16,35.80)(32,40.03)(64,44.22)};
\addlegendentry{Vanilla RL}
\end{axis}
\end{tikzpicture}
\subcaption{More Problems (vs) More Approaches}
\label{fig:prob_vs_approach}
\end{subfigure}
\hfill
\begin{subfigure}[t]{0.49\textwidth}
\centering
\begin{tikzpicture}
\begin{axis}[
    title={Across Six Benchmarks},
    xlabel={$k$},
    ylabel={Avg. Pass@$k$ Accuracy},
    xmode=log,
    log basis x=2,
    xmin=2, xmax=64,
    xtick={2,4,8,16,32,64},
    ymin=19, ymax=45,
    ymajorgrids=true,
    xmajorgrids=true,
    width=\linewidth,
    height=0.68\linewidth,
    legend style={
        draw=none,
        font=\tiny,
        legend columns=1,
        at={(0.98,0.02)},
        anchor=south east,
        row sep=-2pt,
        inner sep=1pt,
        /tikz/every even column/.append style={column sep=3pt}
    },
    legend image post style={scale=0.7},
]
\addplot[color=inkblue, thick, mark=o]
coordinates {(2,22.82)(4,27.83)(8,32.27)(16,36.24)(32,40.05)(64,43.88)};
\addlegendentry{$n=2$}
\addplot[color=teal, thick, mark=o]
coordinates {(2,21.94)(4,26.92)(8,31.49)(16,35.55)(32,39.34)(64,43.29)};
\addlegendentry{$n=4$}
\addplot[color=mutedred, thick, mark=o]
coordinates {(2,20.60)(4,25.36)(8,29.94)(16,34.52)(32,39.02)(64,43.36)};
\addlegendentry{$n=8$}
\addplot[color=gold, thick, mark=o]
coordinates {(2,19.30)(4,24.53)(8,29.27)(16,33.37)(32,36.46)(64,40.57)};
\addlegendentry{$n=16$}
\addplot[
    color=vanillared,
    ultra thick,
    dash pattern=on 6pt off 3pt,
    mark=*,
    mark size=1pt
] coordinates {(2,21.84)(4,26.79)(8,31.43)(16,35.80)(32,40.03)(64,44.22)};
\addlegendentry{Vanilla RL}
\end{axis}
\end{tikzpicture}
\subcaption{Problem-Solving Approaches without Correctness}
\label{fig:incorrect_examples}
\end{subfigure}

\caption{Macro-avg.\ pass@$k$ accuracy after RL across six benchmarks. \textbf{(a)} Comparing mid-trained models on many problems with a single solution \textit{vs.}\ fewer problems with multiple solving approaches. \textbf{(b)} Mid-training with heuristic-guided reasoning chains that have incorrect final answers.}
\label{fig:combined_results}
\end{figure}

\subsection{More Problems or More Approaches?}

To understand the effect of learning multiple problem-solving approaches, we conduct a controlled experiment that fixes the total number of training instances during mid-training while varying the number of unique problems and the number of approaches per problem. More specifically, we ask: \textit{Does learning more problems or more problem-solving approaches during mid-training better benefit RL?} To answer this question, we construct two mid-training datasets using self-generated solutions from the GSM8K training set. In setting 1, we sample a single correct response from the base model for each question, resulting in a dataset of 7{,}408 unique question--solution pairs. In setting 2, we select 463 unique questions and sample 16 distinct problem-solving approaches for each, yielding a total of 7{,}408 training instances. We then perform mid-training using each dataset, followed by RL.

Fig.~\ref{fig:prob_vs_approach} compares the performance of the RL-trained models under the two mid-training regimes. We observe that mid-training with multiple problem-solving approaches per question (Setting~2) consistently outperforms mid-training on a larger number of distinct problems with a single solution each (Setting~1) across all values of $k$. Notably, this improvement persists despite both settings using the same total number of supervised training instances. This result suggests that learning multiple solving approaches, even when applied to a smaller set of problems, provides an avg. relative improvement of $\sim$7\% compared to mid-training on many problems with only one solution each.

\subsection{Problem-Solving Approaches Without Correctness} 
We examine whether exposing model to varied problem-solving approaches alone is sufficient to improve RL, or whether answer correctness is essential. We construct a variant of our mid-training data using procedure in \S~\ref{subsec:midtraining}, but where the generated reasoning chains follow specific Pólya-style approaches yet yield incorrect final answers. This isolates the effect of exposure to different approaches from the effect of training on correct solutions. We mid-train the base model on this data for $n\in\{2,4,8,16\}$ and apply RL with the same setup as in \textsection~\ref{sec:experiments} and compare against vanilla RL training. From Figure~\ref{fig:incorrect_examples}, we observe that when mid-training uses approaches that lead to incorrect answers, increasing $n$ does not improve RL performance. Instead, average pass@$k$ accuracy decreases across all $k$, and these models fall below vanilla RL on all six benchmarks. RL therefore benefits from varied problem-solving approaches only when those approaches lead to correct final answers during mid-training.

\subsection{Can Simple Distillation Produce Diverse Reasoning Approaches?} To investigate whether distillation from a strong reasoning teacher can provide diverse problem-solving approaches, for each question in $\mathcal{D}_{\text{Pólya}}$ we sample 16 reasoning chains from \textbf{QwQ-32B}~\citep{qwq32b}, mid-train a base instruct model on these traces, then apply RL. 

\definecolor{inkblue}{RGB}{56,108,176}
\definecolor{gold}{RGB}{241,196,15}
\definecolor{mutedred}{RGB}{214,69,65}

\providecommand{\dumbbell}[4]{%
  \draw[#1, opacity=0.45, line width=0.8pt]
    (axis cs:#3,#2) -- (axis cs:#4,#2);
  \draw[#1, fill=white, line width=0.7pt]
    (axis cs:#3,#2) circle [radius=1.3pt];
  \draw[#1, fill=#1]
    (axis cs:#4,#2) circle [radius=1.7pt];
}

\begin{wrapfigure}{r}{0.45\textwidth}
\vspace{-0.8\baselineskip}
\centering
\begin{tikzpicture}
\begin{axis}[
  width=\linewidth,
  height=0.88\linewidth,
  xlabel={Score (\%)},
  xlabel style={font=\tiny, yshift=2pt},
  xmin=-4, xmax=92,
  ymin=0.3, ymax=11.5,
  ytick={1.7, 4.5, 7.3, 10.1},
  yticklabels={HMMT 25, AMC 23, AIME 25, AIME 24},
  yticklabel style={font=\tiny},
  xtick={0,20,40,60,80},
  xticklabel style={font=\tiny},
  xmajorgrids=true,
  grid style={dashed, gray!25, line width=0.25pt},
  axis line style={gray!60, line width=0.3pt},
  tick style={gray!50, line width=0.3pt},
  legend style={
    draw=none,
    font=\tiny,
    legend columns=-1,
    at={(0.5,1.02)},
    anchor=south,
    /tikz/every even column/.append style={column sep=4pt},
    inner sep=1pt
  },
  legend image post style={scale=0.6},
]

\addlegendimage{mutedred, mark=*, line width=0.8pt, mark size=1.4pt}
\addlegendentry{Vanilla RL}
\addlegendimage{inkblue, mark=*, line width=0.8pt, mark size=1.4pt}
\addlegendentry{Ours $n{=}16$}
\addlegendimage{gold, mark=*, line width=0.8pt, mark size=1.4pt}
\addlegendentry{Distill+RL}

\dumbbell{mutedred}{10.8}{10.23}{32.90}
\dumbbell{inkblue} {10.1}{ 8.39}{38.60}
\dumbbell{gold}    { 9.4}{ 9.77}{35.54}
\dumbbell{mutedred}{ 8.0}{ 0.53}{16.79}
\dumbbell{inkblue} { 7.3}{ 0.67}{23.34}
\dumbbell{gold}    { 6.6}{ 0.74}{13.58}
\dumbbell{mutedred}{ 5.2}{28.92}{78.18}
\dumbbell{inkblue} { 4.5}{36.24}{84.52}
\dumbbell{gold}    { 3.8}{30.65}{71.90}
\dumbbell{mutedred}{ 2.4}{ 2.00}{15.65}
\dumbbell{inkblue} { 1.7}{ 1.61}{16.73}
\dumbbell{gold}    { 1.0}{ 2.12}{11.96}

\draw[gray!20, dashed, line width=0.25pt] (axis cs:-4, 3.1) -- (axis cs:92, 3.1);
\draw[gray!20, dashed, line width=0.25pt] (axis cs:-4, 5.9) -- (axis cs:92, 5.9);
\draw[gray!20, dashed, line width=0.25pt] (axis cs:-4, 8.7) -- (axis cs:92, 8.7);
\end{axis}
\end{tikzpicture}
\vspace{-0.5\baselineskip}
\caption{Distill+RL vs.\ ours: p@1 (\(\circ\)) and p@64 (\(\bullet\)) across math benchmarks.}
\label{fig:distill_figure}
\vspace{-1.0\baselineskip}
\end{wrapfigure}
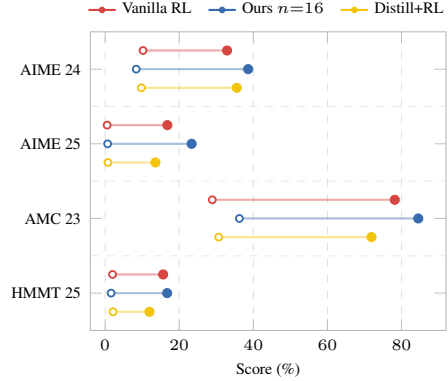 

We compute the Vendi Score (\S~\ref{app:diversity}) on the resulting dataset and compare against our $n=16$ mid-trained RL model in Fig.~\ref{fig:distill_figure}. Fig.~\ref{fig:distill_figure} shows that our mid-trained model achieves better pass@64 after RL and comparable or better pass@1 than the teacher-distillation baseline. The distillation dataset obtains a Vendi Score of \textbf{10.95}, compared to \textbf{13.81} for our mid-training dataset. The distilled model's RL rollouts are also notably more verbose and repetitive, a failure mode also reported by~\citet{li-etal-2025-small-models}. These observations motivate the bootstrapping setup in \S~\ref{sec:experiments} for our investigates, which aims to isolate the effect of learning multiple problem-solving approaches without confounding factors such as capability transfer or distillation efficiency from a stronger teacher.

\subsection{Generalization to Other Domains}
\label{subsec:ood_gen}

Our mid-training relies on Pólya's problem-solving heuristics, which are inherently math-centric. A natural question is whether the resulting problem-solving approaches generalize to other domains. To investigate this, we evaluate on two out-of-domain benchmarks. For coding, we take our mid-trained models (n={32, 64}) from \S~\ref{sec:experiments} and perform RL training on the \textbf{KodCode-Light-RL-10K}~\citep{xu2025kodcode}, then evaluate on HumanEval~\citep{chen2021codex}, comparing against a Vanilla RL (RL applied directly to the base-instruct model without mid-training). Additionally, we take our post-RL models (n={32, 64}) and evaluate on MuSR~\citep{sprague2023musr}, which targets long-context, multi-step reasoning over natural language narratives.

\begin{wraptable}{r}{0.5\textwidth}
\centering
\caption{Results on HumanEval and MuSR.}
\label{tab:ood_results}
\setlength{\tabcolsep}{4pt}
\renewcommand{\arraystretch}{1.15}
\resizebox{\linewidth}{!}{%
\begin{tabular}{@{}lcccc@{}}
\toprule
\multirow{2}{*}{Approach} & \multirow{2}{*}{HumanEval} & \multicolumn{3}{c}{MuSR} \\
\cmidrule(lr){3-5}
 & & Murder & Object & Team \\
 & & Mystery & Placements & Allocations \\
\midrule
Base Instruct & $41.84_{\,0.14}$ & $50.50_{\,0.07}$ & $45.25_{\,0.04}$ & $25.70_{\,0.18}$ \\
Vanilla RL    & $51.14_{\,0.04}$ & $53.15_{\,0.42}$ & $\underline{46.23_{\,0.24}}$ & $23.46_{\,0.14}$ \\
$n{=}64$ (RL) & $\mathbf{52.82_{\,0.09}}$ & $\mathbf{57.36_{\,0.15}}$ & $45.98_{\,0.20}$ & $\underline{38.57_{\,0.05}}$ \\
$n{=}32$ (RL) & $\underline{52.34_{\,0.14}}$ & $\underline{56.94_{\,0.18}}$ & $\mathbf{47.08_{\,0.58}}$ & $\mathbf{39.07_{\,0.44}}$ \\
\bottomrule
\end{tabular}}
\end{wraptable}

From Table~\ref{tab:ood_results}, we observe that our mid-trained models achieve better performance than the Vanilla RL baseline across both out-of-domain benchmarks. On HumanEval, our $n=32$ and $n=64$ models achieve 52.34\% and 52.82\%, respectively, compared to 51.14\% for Vanilla RL. We also observe larger gains on MuSR. On the Murder Mystery sub-task, our models achieve 56.94\% and 57.36\%, compared to 53.15\% for Vanilla RL. In addition to this, on Team Allocations, the Vanilla RL baseline (23.46\%) underperforms even the base instruct model (25.70\%), while our $n=32$ and $n=64$ models achieve \textbf{39.07\%} and \textbf{38.57\%} respectively. These results indicate that the diverse problem-solving approaches induced during mid-training can also generalize to other domains, with the largest gains on tasks that require multi-step reasoning.

\section{Conclusions}
In this paper, we investigated a mid-training strategy to improve RL in LLMs using diverse self-generated data. For each training question, we systematically generated multiple correct solution trajectories guided by George Pólya’s problem-solving heuristics and performed fine-tuning before RL. We showed that this improves the performance of subsequent RL across a range of mathematical reasoning benchmarks, especially at higher pass@$k$ values. We provided a theoretical perspective on how such training benefits RL and can incentivize the composition of diverse problem-solving approaches. Our analysis of reasoning traces provides empirical evidence for this. Furthermore, our analysis reveals that, under a fixed mid-training budget, learning multiple problem-solving approaches per question is more beneficial for RL than increasing the number of distinct training problems with only a single self-generated solution.

\bibliography{main_paper}
\bibliographystyle{plainnat}

\clearpage
\titlecontents{section}[1.5em]
  {\vspace{0.5em}\bfseries\color{blue}}
  {\contentslabel{1.5em}}
  {\hspace*{-1.5em}}
  {\titlerule*[0.5pc]{.}\contentspage}

\titlecontents{subsection}[3.8em]
  {\normalfont\color{blue}}
  {\contentslabel{2.3em}}
  {\hspace*{-2.3em}}
  {\titlerule*[0.5pc]{.}\contentspage}

\vspace{2em}
\noindent{\Large\textbf{\textsf{Table of Contents}}}\par
\vspace{1em}

\startcontents[appendix]
\printcontents[appendix]{}{1}{\setcounter{tocdepth}{2}}
\vspace{2em}

\clearpage
\appendix
\onecolumn
\section{Appendix}

\subsection{Limitations}
 The problem-solving heuristics we use are drawn from P\'olya's \emph{How to Solve It} and are inherently math-centric; while our HumanEval and MuSR results (\S\ref{subsec:ood_gen}) suggest the resulting approaches transfer beyond mathematics, a fully domain-general taxonomy remains an open question. That said, such diverse approaches plausibly already exist  within large-scale pre-training corpora, so our mid-training procedure can be viewed as surfacing and consolidating behaviors the base model has likely been exposed to, rather than introducing fundamentally new ones, a view consistent with the open question raised in \S\ref{subsec:analysisofreasoningtraces} about whether RL gives rise to genuinely emergent behaviors or primarily composes capabilities acquired during pre-training. Although our mid-trained models achieve consistent gains at higher pass@$k$, their pass@$1$ is only slightly better than the vanilla RL and STaR baselines; the focus of our study is not to achieve state-of-the-art single-sample accuracy, but to understand how exposure to multiple problem-solving approaches per question shapes the reasoning behavior elicited by subsequent RL and improve its performance.
 
\subsection{Broader Impact}
Our work contributes to the broader effort of understanding the mechanisms underlying RL training in language models. Rather than proposing RL as a black-box technique for capability gains, we study how the data a model is exposed to prior to RL shapes the behaviors that RL subsequently elicits, and provide both theoretical and empirical evidence that exposure to multiple problem-solving approaches per question encourages RL to compose these approaches within a single reasoning chain. Improved understanding of these mechanisms can inform more principled and predictable post-training pipelines, reduce reliance on trial-and-error scaling of RL compute, and contribute to ongoing discussions about whether RL induces genuinely emergent behaviors or primarily recombines capabilities acquired during pre-training. We do not foresee direct negative societal impacts beyond those generally associated with improved reasoning in language models.

\subsection{Diversity of the Mid-Training Dataset}
\label{app:diversity}
To quantify the diversity of heuristic-guided reasoning traces in our mid-training dataset, we use the \textit{Vendi Score}~\citep{vendi}, a similarity-based metric designed to measure the diversity of a collection. The Vendi Score computes the eigenspectrum of a pairwise similarity matrix and returns the effective number of dissimilar elements in the collection, ranging from 1, when all items are identical, to $n$, the number of items in the collection. We calculate the Vendi Score at the question level by first embedding all $462{,}280$ question--response pairs ($7{,}112$ questions $\times$ $65$ heuristics) using the \texttt{text-embedding-3-small} model~\citep{openai2024embedding}. Since each embedding encodes both the shared question and the heuristic-specific response, we subtract the per-question mean embedding to isolate the variation attributable to the response. We then construct the cosine similarity kernel and compute the Vendi Score as we increase the number of heuristics from 1 to 64. Table~\ref{tab:vendi_results} reports the Vendi Score for different values of $n$, along with the corresponding post-RL performance.

From the Table \ref{tab:vendi_results}, we observe that as $n$ increases, the Vendi Score also increases, indicating that the heuristic-guided data creation mechanism introduces more diverse reasoning approaches. However, the Vendi Score does not reach $n$, as some overlap between problem-solving strategies is expected, such as between analogy-based and auxiliary-problem strategies. We also observe that downstream gains are not strictly monotonic with respect to the Vendi Score, although the top-performing models tend to have higher Vendi Scores. This suggests that learning from more diverse problem-solving approaches can be helpful.

\begin{table}[h]
\caption{Vendi Score and downstream RL performance as the number of problem-solving heuristics $n$ varies.}
\label{tab:vendi_results}
\centering
\small
\begin{tabular}{lccc}
\toprule
\textbf{$n$} & \textbf{Vendi Score} & \textbf{Avg. pass@1} & \textbf{Avg. pass@64} \\
\midrule
Vanilla RL & N/A & 16.87 & 44.22 \\
2  & 1.99  & 17.11 & 43.04 \\
4  & 3.87  & 17.18 & 43.97 \\
8  & 7.44  & 16.75 & 44.09 \\
16 & 13.81 & \textbf{18.35} & \textbf{48.09} \\
32 & 24.16 & 17.32 & 45.21 \\
64 & \textbf{40.56} & 17.91 & 47.62 \\
\bottomrule
\end{tabular}

\end{table}

\subsection{Training Details}
In this section, we provide the details of our training setup. All of experiments are conducted on $4\times$NVIDIA H100 GPUs.
\label{sec:trainingdetails}
\paragraph{Mid-Training \& STaR Setup.} We implement our mid-training using the \texttt{trl}~\citep{trl} framework. All models are optimized with the \textbf{AdamW} optimizer under a cosine learning rate schedule, with the initial learning rate set to $\mathbf{1.0 \times 10^{-6}}$ and a warmup ratio of 0.1. We enable \textbf{bfloat16} mixed-precision training and gradient checkpointing to reduce memory usage, and set the maximum sequence length to \textbf{5190} tokens. To ensure fair comparison across different mid-trained models, we train all configurations for a single epoch with a fixed number of optimization step i.e., \textbf{444 steps}. This is achieved by adjusting the effective batch size according to the number of heuristic variations $n$, as summarized in Table~\ref{tab:effective_batch_size}. For STaR, we first perform inference over $\mathcal{D}{\text{seed}}$ (i.e., the GSM8K training set) with temperature 0 and filter the correct reasoning chains. For questions that are answered incorrectly by the model, we perform rationalization and add the generated reasoning chains to construct $\mathcal{D}{\text{STaR}}$, which contains approximately 7.21k problem–rationale instances.

\begin{table}[h]
\centering
\small
\begin{tabular}{c c}
\toprule
Number of heuristics ($n$) & Effective batch size \\
\midrule
64 & 1024 \\
32 & 512  \\
16 & 256  \\
8  & 128  \\
4  & 64   \\
2  & 32   \\
1  & 16   \\
\bottomrule
\end{tabular}
\caption{Effective batch size used for mid-training with different numbers of heuristic variations.}
\label{tab:effective_batch_size}
\end{table}

\paragraph{Reinforcement Learning.}
We perform RL training with \textbf{DAPO-Math-17k}~\citep{yu2025dapoopensourcellmreinforcement} dataset using the \textbf{GRPO}  algorithm in the \texttt{verl} \citep{verl} framework. To accelerate rollout generation and evaluation, we employ \texttt{vLLM} \citep{vllm}. During training, prompts are truncated to a maximum length of \textbf{1024} tokens and responses are capped at \textbf{3072} tokens. We sample \textbf{16} responses per prompt using a temperature of \textbf{1.0}. Training is conducted with a \textbf{batch size of 32} and a \textbf{mini-batch size of 8}, i.e., one on-policy update and three off-policy updates per optimization step. The model is optimized using the \textbf{AdamW} optimizer with a constant learning rate of $\mathbf{1.0 \times 10^{-6}}$, and gradient checkpointing is enabled to reduce memory usage. We apply asymmetric clipping with $\mathbf{\epsilon_{\text{low}} = 0.2}$ and $\mathbf{\epsilon_{\text{high}} = 0.28}$. Recent work~\citep{scalerl} suggest removing the KL regularization term, as it allows less restrictive model updates while showing no significant evidence of training instability or overoptimization; accordingly, we do not use KL regularization during training. Rewards are based on final answer correctness, and all RL experiments are run for \textbf{1 epoch}.

\subsection{Experiments on Qwen2.5--7B--Instruct}
\label{app:qwen_results}

We conduct additional experiments on \textbf{Qwen2.5--7B--Instruct} model, with the same mid-training procedure as discussed in \textsection~\ref{subsec:midtraining}\&\ref{sec:experiments}. During subsequent RL training, we sample \textbf{8} responses per prompt using a temperature \textbf{1.0}. We discuss the results of our study below.

\renewcommand{\arraystretch}{1.18}
\setlength{\extrarowheight}{0.8pt}
\begin{table*}[htbp!]
\caption{
\textbf{Mid-Training Results.}
Pass@1 and pass@64 performance on four mathematical reasoning benchmarks for Qwen2.5--7B--Instruct.
$\mathcal{D}_{\text{Pólya}}$ denotes mid-training on GSM8K using \emph{self-generated}, diverse, heuristic-guided solution variants, with $n$ variants per question.
Results are reported as mean $\pm$ standard error over three runs.
Best and second-best results are \textbf{bolded} and \underline{underlined}.
}
\label{tab:sft_results_qwen}
\centering
\resizebox{\textwidth}{!}{%
\begin{tabular}{@{}>{\centering\arraybackslash}m{1.8cm}|>{\centering\arraybackslash}m{0.7cm}|cccc!{\vrule}c@{}}
\toprule
\multirow{2}{*}{\textbf{Methods}} &
\multirow{2}{*}{\textit{n}} &
\multicolumn{4}{c!{\vrule}}{\textit{Reasoning Benchmarks} \scriptsize{(pass@1 / pass@64)}} &
\multirow{2}{*}{\textsc{Avg.}} \\
\cmidrule(lr){3-6}
& &
\textsc{AIME 24} &
\textsc{AIME 25} &
\textsc{AMC 23} &
\textsc{HMMT 25} & \\
\midrule
Zero-Shot & N/A &$\best{10.83}\se{0.03}$ / $\best{42.71}\se{1.01}$ &$\best{7.29}\se{0.05}$ / $\best{42.12}\se{0.40}$ &$\best{51.58}\se{0.09}$ / $\second{92.66}\se{0.25}$ &$\best{4.16}\se{0.02}$ / $\second{28.06}\se{0.41}$ &$\best{18.47}$ / $\best{51.39}$ \\
\addlinespace[2pt]
$\mathcal{D}_{\text{STaR}}$ & N/A &$7.70\se{0.01}$ / $37.39\se{0.44}$ &$\second{6.24}\se{0.10}$ / $\second{41.63}\se{0.35}$ &$45.81\se{0.09}$ / $88.00\se{0.36}$ &$2.00\se{0.02}$ / $\best{28.84}\se{0.45}$ &$15.44$ / $48.97$ \\
\midrule
\multirow{4}{*}
{$\mathcal{D}_{\text{Pólya}}$} & 4 &$7.58\se{0.02}$ / $36.46\se{0.31}$ &$4.14\se{0.08}$ / $37.26\se{0.29}$ &$41.46\se{0.06}$ / $91.35\se{0.46}$ &$1.39\se{0.01}$ / $26.02\se{0.72}$ &$13.64$ / $47.77$ \\
\addlinespace[2pt] & 
8 &$9.68\se{0.03}$ / $40.90\se{0.59}$ &$5.91\se{0.04}$ / $41.61\se{0.87}$ &$\second{48.95}\se{0.09}$ / $\best{92.88}\se{0.28}$ &$\second{3.68}\se{0.04}$ / $26.62\se{0.27}$ &$\second{17.06}$ / $\second{50.50}$ \\
\addlinespace[2pt] & 
16 &$9.46\se{0.08}$ / $\second{41.63}\se{0.36}$ &$5.57\se{0.04}$ / $39.59\se{0.34}$ &$48.41\se{0.05}$ / $92.59\se{0.26}$ &$3.60\se{0.03}$ / $26.52\se{0.31}$ &$16.76$ / $50.08$ \\
\addlinespace[2pt] & 
32 &$\second{9.74}\se{0.02}$ / $41.58\se{0.89}$ &$5.76\se{0.04}$ / $39.36\se{0.29}$ &$48.52\se{0.03}$ / $92.48\se{0.32}$ &$3.61\se{0.05}$ / $25.55\se{0.46}$ &$16.91$ / $49.74$ \\
\bottomrule
\end{tabular}%
}
\vspace{0.1cm}
\end{table*}

From Table~\ref{tab:sft_results_qwen}, we observe that the trends in the performance of the mid-trained models differ from those on Llama 3.2--3B--Instruct. On Qwen2.5--7B--Instruct, the zero-shot baseline achieves the highest average pass@1 (\textbf{18.47\%}) and pass@64 (\textbf{51.39\%}), suggesting that this stronger base model already exhibits substantial reasoning diversity. STaR underperforms zero-shot on average (15.44\% / 48.97\%), though it achieves the best pass@64 on HMMT~2025 (28.84\%). Among Pólya configurations, $n=8$ achieves the strongest average performance (\textbf{17.06\%} / \textbf{50.50\%}) and outperforms zero-shot on AMC~2023 pass@64 (92.88\% vs.\ 92.66\%). Increasing $n$ beyond 8 does not yield further gains, with $n=16$ and $n=32$ achieving slightly lower averages (16.76\% / 50.08\% and 16.91\% / 49.74\% respectively).

\renewcommand{\arraystretch}{1.18}
\setlength{\extrarowheight}{0.8pt}

\definecolor{basegray}{RGB}{200,200,200}
\definecolor{vanillared}{RGB}{255,150,150}

\definecolor{v1}{RGB}{68,1,84}
\definecolor{v2}{RGB}{72,40,120}
\definecolor{v3}{RGB}{62,74,137}
\definecolor{v4}{RGB}{38,130,142}
\definecolor{v5}{RGB}{53,183,121}
\definecolor{v6}{RGB}{173,220,48}
\definecolor{v7}{RGB}{253,231,37}

\pgfplotscreateplotcyclelist{customcycle}{
    {v7, thick, mark=o},
    {v6, thick, mark=o},
    {v5, thick, mark=o},
    {v4, thick, mark=o},
}

\begin{figure}[ht]
\centering
\resizebox{\textwidth}{!}{%
\begin{tikzpicture}

\begin{groupplot}[
    group style={group size=2 by 2, horizontal sep=1.2cm, vertical sep=1.2cm},
    cycle list name=customcycle,
    xmode=log,
    log basis x=2,
    xmin=2, xmax=64,
    xtick={2,4,8,16,32,64},
    width=7.2cm,
    height=4.3cm,
    ymajorgrids=true,
    xmajorgrids=true
]

\nextgroupplot[
    title={AIME 24},
    ylabel={Pass@k},
    legend to name=commonlegend,
    legend style={draw=none, font=\small, legend columns=6, column sep=0.5em},
    ymin=16, ymax=50
]

\addplot coordinates {(2,17.75)(4,22.30)(8,26.76)(16,32.02)(32,38.00)(64,43.80)};
\addplot coordinates {(2,17.02)(4,20.69)(8,23.96)(16,27.85)(32,32.48)(64,37.94)};
\addplot coordinates {(2,18.39)(4,23.39)(8,28.57)(16,34.43)(32,41.41)(64,48.68)};
\addplot coordinates {(2,16.17)(4,20.07)(8,23.96)(16,27.92)(32,32.01)(64,36.65)};

\addplot[vanillared, ultra thick, dash pattern=on 6pt off 3pt, mark=*, mark size=1pt, forget plot]
    coordinates {(2,15.43)(4,18.49)(8,21.58)(16,24.82)(32,28.66)(64,33.66)};
\addplot[basegray, thick, mark=*, forget plot]
    coordinates {(2,15.76)(4,19.45)(8,23.24)(16,27.74)(32,33.08)(64,39.06)};

\addlegendentry{$n=32$}
\addlegendentry{$n=16$}
\addlegendentry{$n=8$}
\addlegendentry{$n=4$}
\addlegendentry{Vanilla RL}
\addlegendentry{STaR+RL}

\addplot[
    color=v5,
    ultra thick,
    mark=o,
    on layer=foreground,
    forget plot
] coordinates {(2,18.39)(4,23.39)(8,28.57)(16,34.43)(32,41.41)(64,48.68)};

\nextgroupplot[title={AIME 25}, ymin=11, ymax=43]

\addplot coordinates {(2,16.49)(4,21.49)(8,25.91)(16,30.15)(32,35.20)(64,41.52)};
\addplot coordinates {(2,12.35)(4,16.72)(8,21.62)(16,27.10)(32,32.94)(64,38.78)};
\addplot coordinates {(2,12.85)(4,17.16)(8,21.73)(16,27.11)(32,33.11)(64,39.17)};
\addplot coordinates {(2,12.11)(4,16.87)(8,22.67)(16,29.31)(32,36.16)(64,42.54)};

\addplot[vanillared, ultra thick, dash pattern=on 6pt off 3pt, mark=*, mark size=1pt, forget plot]
    coordinates {(2,15.72)(4,20.87)(8,25.53)(16,29.53)(32,33.72)(64,38.56)};
\addplot[basegray, thick, mark=*, forget plot]
    coordinates {(2,15.07)(4,20.69)(8,26.13)(16,31.05)(32,35.26)(64,39.23)};

\addplot[
    color=v7,
    ultra thick,
    mark=o,
    on layer=foreground,
    forget plot
] coordinates {(2,16.49)(4,21.49)(8,25.91)(16,30.15)(32,35.20)(64,41.52)};

\nextgroupplot[title={AMC 23}, ylabel={Pass@k}, ymin=59, ymax=94]

\addplot coordinates {(2,68.32)(4,77.35)(8,83.54)(16,86.92)(32,88.69)(64,90.76)};
\addplot coordinates {(2,68.56)(4,76.07)(8,82.17)(16,86.78)(32,89.94)(64,92.32)};
\addplot coordinates {(2,68.72)(4,76.06)(8,82.12)(16,86.89)(32,90.20)(64,92.59)};
\addplot coordinates {(2,60.20)(4,67.44)(8,73.85)(16,79.23)(32,83.55)(64,87.76)};

\addplot[vanillared, ultra thick, dash pattern=on 6pt off 3pt, mark=*, mark size=1pt, forget plot]
    coordinates {(2,62.64)(4,71.23)(8,78.57)(16,84.56)(32,88.76)(64,91.14)};
\addplot[basegray, thick, mark=*, forget plot]
    coordinates {(2,61.28)(4,69.61)(8,76.12)(16,81.82)(32,87.27)(64,92.32)};

\addplot[
    color=v5,
    ultra thick,
    mark=o,
    on layer=foreground,
    forget plot
] coordinates {(2,68.72)(4,76.06)(8,82.12)(16,86.89)(32,90.20)(64,92.59)};

\nextgroupplot[title={HMMT 25}, ymin=2, ymax=31]

\addplot coordinates {(2,6.75)(4,9.15)(8,11.51)(16,14.05)(32,17.32)(64,21.73)};
\addplot coordinates {(2,7.58)(4,10.64)(8,13.74)(16,16.79)(32,20.05)(64,24.02)};
\addplot coordinates {(2,8.29)(4,11.72)(8,15.16)(16,18.81)(32,22.77)(64,26.87)};
\addplot coordinates {(2,2.06)(4,3.89)(8,7.01)(16,11.58)(32,17.05)(64,22.52)};

\addplot[vanillared, ultra thick, dash pattern=on 6pt off 3pt, mark=*, mark size=1pt, forget plot]
    coordinates {(2,7.44)(4,10.90)(8,14.57)(16,18.54)(32,22.98)(64,27.62)};
\addplot[basegray, thick, mark=*, forget plot]
    coordinates {(2,2.75)(4,5.16)(8,9.17)(16,15.01)(32,22.26)(64,29.92)};

\addplot[
    color=v5,
    ultra thick,
    mark=o,
    on layer=foreground,
    forget plot
] coordinates {(2,8.29)(4,11.72)(8,15.16)(16,18.81)(32,22.77)(64,26.87)};

\end{groupplot}

\node[anchor=north, yshift=-0.3em, font=\small] at (current bounding box.south) {
    \begin{tabular}{c}
        \pgfplotslegendfromname{commonlegend} \\[0.1em]
        \tikz{
            \draw[vanillared, ultra thick, dash pattern=on 6pt off 3pt] (0,0) -- (0.5,0)
                node[circle, fill=vanillared, inner sep=1pt, pos=0.5] {};
            \node[right] at (0.55,0) {Vanilla RL};
            \draw[basegray, thick] (2.8,0) -- (3.3,0)
                node[circle, fill=basegray, inner sep=1.5pt, pos=0.5] {};
            \node[right] at (3.35,0) {STaR+RL};
        }
    \end{tabular}
};

\end{tikzpicture}
}
\caption{Pass@\(k\) performance of RL-trained models on four mathematical reasoning benchmarks. The x-axis shows the number of samples \(k\) (log scale), and the y-axis shows accuracy. Each curve corresponds to a model first mid-trained with \(n \in \{4,8,16,32\}\) distinct Pólya-style problem-solving heuristics and subsequently trained with reinforcement learning. Highlighted curves indicate the best-performing \(n\) for each benchmark.}
\label{fig:pass_at_k_qwen}
\end{figure}
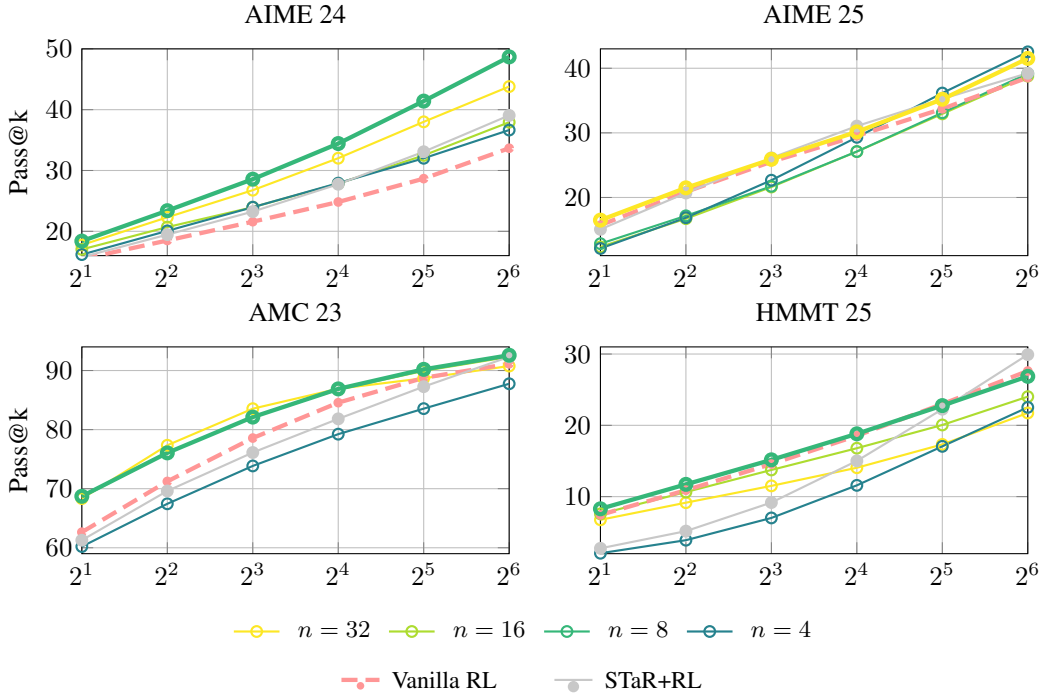

From Fig.~\ref{fig:pass_at_k_qwen}, we observe trends similar to those on Llama 3.2--3B--Instruct: mid-trained models generally outperform the Vanilla RL and STaR+RL baselines across benchmarks. Among Pólya configurations, $n=8$ achieves the strongest performance on three of four benchmarks (AIME~24, AMC~23, and HMMT~25), with $n=4$ leading on AIME~25 at pass@64 (42.54\%). On AIME~24 and AMC~23, $n=8$ outperforms both baselines across all values of $k$, reaching pass@64 values of \textbf{48.68\%} and \textbf{92.59\%} respectively. On HMMT~25, $n=8$ dominates at lower $k$ but the baselines partially catch up at pass@64 (Vanilla RL: 27.62\%, STaR+RL: 29.92\%, vs.\ $n=8$: 26.87\%). These results corroborate our rollout-matching hypothesis from Section~\ref{sec:experiments}: with rollout size $g=8$, the best-performing $n$ matches $g$ on three of four benchmarks. Overall, the trends on Qwen2.5--7B--Instruct are consistent with our main results that mid-training with diverse self-generated data improves subsequent RL training.

\subsection{Proofs and Derivations}
\label{sec:proofs}
In this section, we provide the formal derivation for Theorem 4.1 and Proposition 4.2. We begin by deriving the exact analytical form of the first-order probability mass shift, and then specialize this result to the uni-modal and $N$-modal regimes.

\subsubsection{Preliminaries}
Let $\pi_{\theta}(\cdot \mid x)$ denote a language model parameterized by $\theta$, which defines a categorical distribution over tokens via a softmax over logits $\{z_j\}$. Given a context $x$, a token $y_t \sim \pi_{\theta}(\cdot \mid x)$ is sampled from this distribution.  

We consider a single policy-gradient update with learning rate $\eta$ and scalar advantage $A$, driven by the gradient $\nabla_{\theta} \log \pi_{\theta}(y_t \mid x)$. The effective change to the logit $z_j$ induced by this update is
\begin{equation}
\Delta z_j
= \eta \cdot A \cdot \frac{\partial \log \pi_{\theta}(y_t \mid x)}{\partial z_j}.
\end{equation}

Since $\pi_{\theta}(\cdot \mid x)$ is a softmax policy, the gradient of the log-probability with respect to the logits can be written as
\begin{equation}
\frac{\partial \log \pi_{\theta}(y_t \mid x)}{\partial z_j}
= \mathbbm{1}_{\{j = y_t\}} - \pi_{\theta}(j \mid x).
\end{equation}

Substituting, we obtain
\begin{equation}
\Delta z_j
= \eta \cdot A \cdot \left( \mathbbm{1}_{\{j=y_t\}} - \pi_{\theta}(j \mid x) \right).
\label{eq:logit-change}
\end{equation}

\subsubsection{Derivation of First-Order Probability Change}

\begin{proof}[Proof of Theorem 4.1]
Let $\pi_\theta(\cdot\mid x)$ be the softmax policy model and let $y_t\sim \pi_\theta(\cdot\mid x)$ be the sampled token. A single policy-gradient update with learning rate $\eta$ and advantage $A$ induces the logit change given in \eqref{eq:logit-change}.

The first-order change in the probability of the sampled token $y_t$ is obtained by the Taylor expansion:
\begin{equation}
\Delta \pi_\theta(y_t\mid x)
\approx
\sum_{j\in\mathcal V}
\frac{\partial \pi_\theta(y_t\mid x)}{\partial z_j}\,\Delta z_j.
\label{eq:first-order-change}
\end{equation}

For a softmax policy, the Jacobian with respect to the logits is
\begin{equation}
\frac{\partial \pi_\theta(y_t\mid x)}{\partial z_j}
=
\pi_\theta(y_t\mid x)
\left(
\mathbbm{1}_{\{j=y_t\}}-\pi_\theta(j\mid x)
\right).
\label{eq:jacobian}
\end{equation}

Substituting the expressions for $\Delta z_j$ (equation \ref{eq:logit-change}) and the Jacobian (equation \ref{eq:jacobian}) to Equation \ref{eq:first-order-change}: 
\[
\begin{aligned}
\Delta \pi_\theta(y_t\mid x)
&\approx
\sum_{j\in\mathcal V}
\pi_\theta(y_t\mid x)
\left(
\mathbbm{1}_{\{j=y_t\}}-\pi_\theta(j\mid x)
\right)
\cdot
\eta A
\left(
\mathbbm{1}_{\{j=y_t\}}-\pi_\theta(j\mid x)
\right)
\\
&=
\eta A\,\pi_\theta(y_t\mid x)
\sum_{j\in\mathcal V}
\left(
\mathbbm{1}_{\{j=y_t\}}-\pi_\theta(j\mid x)
\right)^2.
\end{aligned}
\]

We now decompose the sum by separating the $j = y_t$ term from the rest:
\[
\sum_{j\in\mathcal V}
\left(
\mathbbm{1}_{\{j=y_t\}}-\pi_\theta(j\mid x)
\right)^2
=
\left(1 - \pi_\theta(y_t\mid x)\right)^2
+
\sum_{j\neq y_t}
\left(0 - \pi_\theta(j\mid x)\right)^2.
\]

Simplifying the second term:
\[
\sum_{j\neq y_t}
\left(0 - \pi_\theta(j\mid x)\right)^2
=
\sum_{j\neq y_t}
\pi_\theta(j\mid x)^2.
\]

Therefore, we obtain the general formula:
\begin{equation}
\Delta \pi_\theta(y_t\mid x)
=
\eta A\,\pi_\theta(y_t\mid x)
\Bigl[
(1-\pi_\theta(y_t\mid x))^2
+
\sum_{j\neq y_t}\pi_\theta(j\mid x)^2
\Bigr].
\label{eq:delta-pi}  
\end{equation}

This completes the derivation of the general first-order probability change.

We now analyze two important special cases of \eqref{eq:delta-pi}.

\paragraph{Case (i): Uni-modal regime.}
Assume $\pi_\theta(y_t\mid x)=1-\epsilon$ with residual probability mass $\epsilon$, where $\epsilon \ll 1$. 

The first term in \eqref{eq:delta-pi} becomes:
\[
(1-\pi_\theta(y_t\mid x))^2 = (1-(1-\epsilon))^2 = \epsilon^2.
\]

For the second term of equation \ref{eq:delta-pi}, note that since probabilities are non-negative and sum to one, we have $\sum_{j\neq y_t}\pi_\theta(j\mid x) = \epsilon$. Since, probability of any token is non-negative the following inequality holds:

\[
\sum_{j\neq y_t}\pi_\theta(j\mid x)^2
\le
\left(\sum_{j\neq y_t}\pi_\theta(j\mid x)\right)
\max_{j\neq y_t} \pi_\theta(j\mid x)
\le
\epsilon \cdot \epsilon
=
\epsilon^2.
\]

In fact, if the residual mass is distributed over multiple tokens, we have $\sum_{j\neq y_t}\pi_\theta(j\mid x)^2 = O(\epsilon^2)$ with a constant factor less than 1.

Substituting into \eqref{eq:delta-pi}:
\[
\Delta \pi_\theta(y_t\mid x)
=
\eta A(1-\epsilon)\bigl[\epsilon^2+O(\epsilon^2)\bigr].
\]

Expanding $(1-\epsilon)[\epsilon^2 + O(\epsilon^2)]$:
\[
(1-\epsilon)[\epsilon^2 + O(\epsilon^2)]
=
\epsilon^2 + O(\epsilon^2) - \epsilon^3 - O(\epsilon^3)
=
\epsilon^2 + O(\epsilon^2),
\]
where the $O(\epsilon^2)$ term absorbs the $-\epsilon^3$ contribution for small $\epsilon$.

Therefore:
\[
\Delta \pi_\theta(y_t\mid x)
\approx
\eta A\,\epsilon^2.
\]

\paragraph{Case (ii): $N$-modal regime.}
Assume the probability mass $1-\epsilon$ is evenly split among $N$ dominant modes, so that each mode has probability
\[
\pi_\theta(j\mid x)\approx\frac{1-\epsilon}{N}
\quad \text{for } j \in \{1, \ldots, N\}.
\]

In particular, for the sampled token $y_t$ (which is one of these $N$ modes):
\[
\pi_\theta(y_t\mid x)\approx\frac{1-\epsilon}{N}.
\]

We compute the sum of squared probabilities. The $N$ dominant modes contribute:
\[
N\left(\frac{1-\epsilon}{N}\right)^2
=
N \cdot \frac{(1-\epsilon)^2}{N^2}
=
\frac{(1-\epsilon)^2}{N}.
\]

The residual mass $\epsilon$ contributes at most $O(\epsilon^2)$ (by the same argument as in the uni-modal case). Therefore:
\[
\sum_{j\in\mathcal V}\pi_\theta(j\mid x)^2
=
\frac{(1-\epsilon)^2}{N}
+
O(\epsilon^2).
\]

Now we evaluate the bracketed expression in \eqref{eq:delta-pi}. The first term is:
\[
(1-\pi_\theta(y_t\mid x))^2
=
\left(1-\frac{1-\epsilon}{N}\right)^2
=
\left(\frac{N - (1-\epsilon)}{N}\right)^2
=
\left(\frac{N-1+\epsilon}{N}\right)^2.
\]

Expanding:
\begin{equation}
\left(\frac{N-1+\epsilon}{N}\right)^2
=
\frac{(N-1)^2 + 2(N-1)\epsilon + \epsilon^2}{N^2}
=
\frac{(N-1)^2}{N^2} + O\left(\frac{\epsilon}{N}\right).
\label{eq:first-term}    
\end{equation}

The second term is:
\[
\sum_{j\neq y_t}\pi_\theta(j\mid x)^2
=
\sum_{j\in\mathcal V}\pi_\theta(j\mid x)^2 - \pi_\theta(y_t\mid x)^2
=
\frac{(1-\epsilon)^2}{N} - \left(\frac{1-\epsilon}{N}\right)^2 + O(\epsilon^2).
\]

Simplifying:
\[
\frac{(1-\epsilon)^2}{N} - \frac{(1-\epsilon)^2}{N^2}
=
\frac{(1-\epsilon)^2}{N}\left(1 - \frac{1}{N}\right)
=
\frac{(1-\epsilon)^2(N-1)}{N^2}.
\]

Expanding $(1-\epsilon)^2 = 1 - 2\epsilon + \epsilon^2$:
\begin{equation}
\frac{(N-1)(1-2\epsilon+\epsilon^2)}{N^2}
=
\frac{N-1}{N^2} - \frac{2(N-1)\epsilon}{N^2} + O\left(\frac{\epsilon^2}{N}\right)
=
\frac{N-1}{N^2} + O\left(\frac{\epsilon}{N}\right).
\label{eq:second-term}    
\end{equation}

Adding the two terms (equation \ref{eq:first-term} and \ref{eq:second-term}):
\[
\frac{(N-1)^2}{N^2} + O\left(\frac{\epsilon}{N}\right) + \frac{N-1}{N^2} + O\left(\frac{\epsilon}{N}\right)
=
\frac{(N-1)^2 + (N-1)}{N^2} + O\left(\frac{\epsilon}{N}\right).
\]

Factoring the numerator:
\[
(N-1)^2 + (N-1) = (N-1)(N-1+1) = (N-1)N.
\]

Therefore:
\[
\frac{(N-1)N}{N^2} + O\left(\frac{\epsilon}{N}\right)
=
\frac{N-1}{N} + O\left(\frac{\epsilon}{N}\right)
=
1 - \frac{1}{N} + O\left(\frac{\epsilon}{N}\right).
\]

Substituting into \eqref{eq:delta-pi}:
\[
\Delta \pi_\theta(y_t\mid x)
=
\eta A\,\frac{1-\epsilon}{N}
\left[1 - \frac{1}{N} + O\left(\frac{\epsilon}{N}\right)\right].
\]

Expanding:
\[
\eta A\,\frac{1-\epsilon}{N}
\left(1 - \frac{1}{N}\right)
+ \eta A\,\frac{1-\epsilon}{N} \cdot O\left(\frac{\epsilon}{N}\right)
=
\eta A\,\frac{1-\epsilon}{N}
\left(1 - \frac{1}{N}\right)
+
O\!\left(\frac{\epsilon}{N^2}\right).
\]

Factoring out:
\[
\eta A\,\frac{1}{N}\left(1-\frac{1}{N}\right)(1-\epsilon)
+
O\!\left(\frac{\epsilon}{N^2}\right).
\]

Since $(1-\epsilon) = 1 + O(\epsilon)$, we can absorb this into the error term:
\[
\Delta \pi_\theta(y_t\mid x)
=
\eta A\,\frac{1}{N}\left(1-\frac{1}{N}\right)
+
O\!\left(\frac{\epsilon}{N}\right).
\]

This completes the proof.
\end{proof}

\subsubsection{Redistribution of probability mass in the $N$-Modal Regime}
\paragraph{Proposition 3.2}
Consider the $N$-modal regime in which probability mass $1-\epsilon$ is distributed among $N \ll |\mathcal{V}|$ dominant modes. Under a single policy-gradient update with negative advantage $A<0$, the probability mass removed from the sampled token $y_t$ is redistributed predominantly to the remaining $N-1$ dominant modes.

\begin{proof}
Let $y_t$ be the sampled token with negative advantage $A$. We compute the probability update for any other token $j \neq y_t$. Starting from the Jacobian expansion:
\[
\Delta \pi_\theta(j\mid x)
\approx
\sum_{k\in\mathcal V}
\frac{\partial \pi_\theta(j\mid x)}{\partial z_k}\,\Delta z_k.
\]

Using the softmax Jacobian and logit update:
\[
\Delta \pi_\theta(j\mid x)
\approx
\eta A \sum_{k\in\mathcal V}
\pi_\theta(j\mid x)
\left(\mathbbm{1}_{\{k=j\}}-\pi_\theta(k\mid x)\right)
\left(\mathbbm{1}_{\{k=y_t\}}-\pi_\theta(k\mid x)\right).
\]

Expanding the product inside the summation:
\begin{align*}
&(\mathbbm{1}_{\{k=j\}}-\pi_\theta(k\mid x))(\mathbbm{1}_{\{k=y_t\}}-\pi_\theta(k\mid x)) \\
&\quad = \mathbbm{1}_{\{k=j\}}\mathbbm{1}_{\{k=y_t\}} - \mathbbm{1}_{\{k=j\}}\pi_\theta(k\mid x) - \mathbbm{1}_{\{k=y_t\}}\pi_\theta(k\mid x) + \pi_\theta(k\mid x)^2.
\end{align*}

Summing over $k$ (noting that $\mathbbm{1}_{\{k=j\}}\mathbbm{1}_{\{k=y_t\}} = 0$ since $j \neq y_t$):
\begin{multline*}
\sum_{k\in\mathcal V} \left[\mathbbm{1}_{\{k=j\}}\mathbbm{1}_{\{k=y_t\}} - \mathbbm{1}_{\{k=j\}}\pi_\theta(k\mid x) - \mathbbm{1}_{\{k=y_t\}}\pi_\theta(k\mid x) + \pi_\theta(k\mid x)^2\right] \\
= 0 - \pi_\theta(j\mid x) - \pi_\theta(y_t\mid x) + \sum_{k\in\mathcal V} \pi_\theta(k\mid x)^2.
\end{multline*}

Let $S = \sum_{k} \pi_\theta(k\mid x)^2$. The update expression becomes:
\begin{equation}
\label{eq:exact_j_update}
\Delta \pi_\theta(j\mid x) \approx \eta A \, \pi_\theta(j\mid x) \left[ S - \pi_\theta(j\mid x) - \pi_\theta(y_t\mid x) \right].
\end{equation}

We now apply the $N$-modal assumption. For dominant modes, $\pi_\theta(k\mid x) \approx \frac{1-\epsilon}{N}$ for $k \in \mathcal{M}_{dom}$. The sum of squared probabilities is dominated by the $N$ modes:
\[
S = \sum_{k\in\mathcal V} \pi_\theta(k\mid x)^2 \approx N \left(\frac{1-\epsilon}{N}\right)^2 + O(\epsilon^2) = \frac{(1-\epsilon)^2}{N} + O(\epsilon^2).
\]

\textbf{Case (i): Update for another dominant mode ($j \in \mathcal{M}_{dom}, j \neq y_t$):}

Under our assumption that all dominant modes have similar probability, we substitute $\pi_\theta(j\mid x) \approx \pi_\theta(y_t\mid x) \approx \frac{1-\epsilon}{N}$ into \eqref{eq:exact_j_update}:
\[
\Delta \pi_{dom} \approx \eta A \cdot \frac{1-\epsilon}{N} \left[ \frac{(1-\epsilon)^2}{N} - \frac{1-\epsilon}{N} - \frac{1-\epsilon}{N} \right].
\]

Simplifying the bracketed term:
\[
\frac{(1-\epsilon)^2}{N} - \frac{2(1-\epsilon)}{N} 
= \frac{1}{N}\left[(1-\epsilon)^2 - 2(1-\epsilon)\right]
= \frac{1-\epsilon}{N}\left[(1-\epsilon) - 2\right]
= -\frac{(1-\epsilon)(1+\epsilon)}{N}.
\]

Therefore:
\[
\Delta \pi_{dom} = \eta A \cdot \frac{1-\epsilon}{N} \cdot \left(-\frac{(1-\epsilon)(1+\epsilon)}{N}\right)
= -\eta A \cdot \frac{(1-\epsilon)^2(1+\epsilon)}{N^2}.
\]

Since $A < 0$, write $A = -|A|$:
\[
\Delta \pi_{dom} = \frac{\eta |A|(1-\epsilon)^2(1+\epsilon)}{N^2}.
\]

The total mass gained by all $N-1$ dominant neighbors is:
\[
\Delta \Pi_{dom} = (N-1) \cdot \frac{\eta |A|(1-\epsilon)^2(1+\epsilon)}{N^2} = \frac{\eta |A|(N-1)(1-\epsilon)^2(1+\epsilon)}{N^2}.
\]

\textbf{Case (ii): Update for a tail/non-dominant token ($j \in \mathcal{T}$):}

For tail tokens, $\pi_\theta(j\mid x) = O(\epsilon/|\mathcal{V}|)$ is very small. The bracket term in \eqref{eq:exact_j_update} becomes:
\[
S - \pi_{tail} - \pi_{y_t} \approx \frac{(1-\epsilon)^2}{N} - 0 - \frac{1-\epsilon}{N} = \frac{(1-\epsilon)}{N}\left[(1-\epsilon) - 1\right] = -\frac{\epsilon(1-\epsilon)}{N}.
\]

Therefore:
\[
\Delta \pi_{tail} = \eta A \cdot \pi_{tail} \cdot \left(-\frac{\epsilon(1-\epsilon)}{N}\right) = \frac{\eta |A| \epsilon(1-\epsilon)}{N} \cdot \pi_{tail}.
\]

Since $\pi_{tail} = O(\epsilon/|\mathcal{V}|)$, this gives:
\[
\Delta \pi_{tail} = O\left(\frac{\eta |A| \epsilon^2}{N|\mathcal{V}|}\right),
\]
which is negligible compared to $\Delta \pi_{dom} = O(\eta |A|/N^2)$ when $\epsilon \ll 1$ and $|\mathcal{V}|$ is large.

From Theorem 3.1, the mass lost by $y_t$ is:
\[
\Delta \pi_\theta(y_t\mid x) = \eta A \cdot \frac{1-\epsilon}{N}\left(1-\frac{1}{N}\right) + O\left(\frac{\epsilon}{N}\right) = -\frac{\eta |A|(1-\epsilon)(N-1)}{N^2} + O\left(\frac{\epsilon}{N}\right).
\]

The mass gained by the $N-1$ dominant neighbors is:
\[
\Delta \Pi_{dom} = \frac{\eta |A|(N-1)(1-\epsilon)^2(1+\epsilon)}{N^2}.
\]

Expanding $(1-\epsilon)^2(1+\epsilon) = (1-2\epsilon+\epsilon^2)(1+\epsilon) = 1 - \epsilon - \epsilon^2 + O(\epsilon^3)$, we have:
\[
\Delta \Pi_{dom} = \frac{\eta |A|(N-1)(1-\epsilon)}{N^2} + O\left(\frac{\epsilon}{N}\right),
\]
which matches the mass lost from $y_t$ to leading order.

The mass redistribution is dominated by the $N-1$ dominant modes, which collectively gain $\frac{\eta|A|(N-1)(1-\epsilon)^2(1+\epsilon)}{N^2}$ mass. The tail tokens receive only $O(\epsilon^2/(N|\mathcal{V}|))$ mass per token, which is negligible. Thus, probability mass is redistributed almost entirely among the dominant modes.
\end{proof}

\clearpage
\subsection{Prompts}
\label{subsection:prompts}

\subsubsection{Data Generation}
\label{subsubsection:data_generation}
\begin{center}
\begin{tcolorbox}[
    title=Prompt for \textit{heuristic-specific data generation},
    breakable,  
    top=1mm,
    bottom=1mm,
    left=1mm,
    right=1mm
]
\ttfamily\scriptsize
\textbf{System:} You are a helpful assistant that generates the reasoning chain by strictly following the heuristic instructions to answer the given question. There should be no mention of you being instructed to follow the heuristic instructions in your generated reasoning chain. YOU WILL NOT MENTION OR REFER TO THE HEURISTIC INSTRUCTIONS IN YOUR GENERATED REASONING CHAIN.\\ \\
\textbf{[Few-shot example 1]}\\
\textbf{User:} \\
\texttt{<question>}\\
\{question\_1\}. provide your final answer inside \textbackslash{}boxed\{\} notation\\
\texttt{</question>}\\
\texttt{<heuristic\_instructions>}\\
\{heuristic\_description\}\\
\texttt{</heuristic\_instructions>}\\
Only output the reasoning chain. No other comments or references to the heuristic instructions in your output!\\ \\
\textbf{Assistant:} \{example\_reasoning\_chain\_1\}\\ \\
\textbf{[Few-shot example 2]}\\
\textbf{User:} \\
\texttt{<question>}\\
\{question\_2\}. provide your final answer inside \textbackslash{}boxed\{\} notation\\
\texttt{</question>}\\
\texttt{<heuristic\_instructions>}\\
\{heuristic\_description\}\\
\texttt{</heuristic\_instructions>}\\
Only output the reasoning chain. No other comments or references to the heuristic instructions in your output!\\ \\
\textbf{Assistant:} \{example\_reasoning\_chain\_2\}\\ \\
\textit{... (2 more few-shot examples) ...}\\ \\
\textbf{[Target generation]}\\
\textbf{User:} \\
\texttt{<question>}\\
\{target\_question\}. provide your final answer inside \textbackslash{}boxed\{\} notation\\
\texttt{</question>}\\
\texttt{<heuristic\_instructions>}\\
\{heuristic\_description\}\\
\texttt{</heuristic\_instructions>}\\
Only output the reasoning chain. No other comments or references to the heuristic instructions in your output!
\end{tcolorbox}
\captionof{figure}{Prompt template for heuristic-specific data generation. For each heuristic $h$, we include its description $D_h$ and 4 few-shot exemplars $\mathcal{E}_h$ demonstrating the heuristic applied to different questions. The model samples 128 responses per question--heuristic pair.}
\label{fig:datagen_prompt}
\end{center}
\begin{center}
\begin{tcolorbox}[
    title=\textit{Reward Model Scoring for Heuristic Adherence},
    breakable,  
    top=1mm,
    bottom=1mm,
    left=1mm,
    right=1mm
]
\ttfamily\scriptsize
\textbf{[Model]}\\
Skywork-Reward-V2-Llama-3.2-3B\\ \\
\textbf{[Input Format]}\\
\texttt{[User]:}\\
\{question\}\\
\\
Solve this problem using the following approach:\\
\texttt{<heuristic>}\\
\{heuristic\_description\}\\
\texttt{</heuristic>}\\ \\
\texttt{[Assistant]:}\\
\{generated\_reasoning\_chain\}\\ \\
\textbf{[Output]}\\
Scalar reward score $\in \mathbb{R}$\\ \\
\textbf{[Selection]}\\
For each (question, heuristic) pair with $k$ candidate responses:\\
\hspace{1em}1. Filter candidates by answer correctness via \texttt{Math-Verify}\\
\hspace{1em}2. Score remaining candidates with reward model\\
\hspace{1em}3. Select response with highest score as final heuristic-specific solution
\end{tcolorbox}
\captionof{figure}{Reward model scoring pipeline. After filtering for correctness, Skywork-Reward-V2 ranks candidate responses by heuristic adherence. The highest-scoring response per (question, heuristic) pair is selected for mid-training.}
\label{fig:reward_scoring}
\end{center}

\subsubsection{Reinforcement Learning}
\label{subsubsection:rl_prompt}
\begin{center}
\begin{tcolorbox}[
    title=\textit{RL Training Data Format},
    top=1mm,
    bottom=1mm,
    left=1mm,
    right=1mm
]
\ttfamily\scriptsize
\textbf{Input:} \{question\}\\ \\
\textbf{Formatted Prompt:}\\
\texttt{[User]:} \{question\} Let's think step by step and output the final answer within \textbackslash boxed\{\}.\\ \\
\end{tcolorbox}
\captionof{figure}{RL training prompt format. Each question is paired with a chain-of-thought instruction; correctness is verified programmatically.}
\label{fig:rl_prompt_format}
\end{center}

\subsubsection{Compositional Analysis}
\label{subsubsection:emergence}
For each heuristic classification, we construct the following prompt:
\begin{center}
\begin{tcolorbox}[
    title=Prompt for \textit{heuristic classification},
    breakable,  
    top=1mm,
    bottom=1mm,
    left=1mm,
    right=1mm
]
\ttfamily\scriptsize
\textbf{[System prompt]}\\
You are a classifier that determines whether a mathematical reasoning chain exhibits a specific problem-solving heuristic.\\
\\
First, provide brief reasoning about whether the heuristic is present.\\
Then, give your final decision as exactly "Decision: Yes" or "Decision: No".\\
\\
Format:\\
Reasoning: <your analysis>\\
Decision: <Yes or No>\\ \\
\textbf{[Few-shot example 1 - Positive]}\\ 
Heuristic: \{heuristic\_name\}\\
Description: \{heuristic\_description\}\\
\\
Reasoning chain:\\
\{positive\_example\_1\}\\
\\
Does this exhibit the '\{heuristic\_name\}' heuristic?\\
\textbf{Assistant:} Reasoning: The reasoning chain demonstrates this heuristic.\\
Decision: Yes\\ \\
\textbf{[Few-shot example 2 - Positive]}\\
\textit{(Same format as above with positive\_example\_2)}\\ \\
\textbf{[Few-shot example 3 - Negative]}\\
Heuristic: \{heuristic\_name\}\\
Description: \{heuristic\_description\}\\
\\
Reasoning chain:\\
\{negative\_example\_from\_other\_heuristic\}\\
\\
Does this exhibit the '\{heuristic\_name\}' heuristic?\\
\textbf{Assistant:} Reasoning: The reasoning chain does not exhibit this heuristic.\\
Decision: No\\ \\
\textbf{[Few-shot example 4 - Negative]}\\
\textit{(Same format as above with another negative example)}\\ \\
\textbf{[Target classification]}\\
Heuristic: \{heuristic\_name\}\\
Description: \{heuristic\_description\}\\
\\
Reasoning chain:\\
\{target\_response\}\\
\\
Does this exhibit the '\{heuristic\_name\}' heuristic?
\end{tcolorbox}
\captionof{figure}{Prompt template for LLM-as-judge heuristic classification using GPT-4o-mini. Each model response is evaluated against all 64 heuristics independently. The model outputs reasoning followed by a binary decision.}
\label{fig:heuristic_prompt}
\end{center}


\subsection{Evaluation Datasets}
\label{appendix:eval_data}
All reported results are specific to each dataset's respective test set. 
\begin{table}[htbp!]
\centering
\begin{tabular}{@{}lr@{}}
\toprule
\textbf{Dataset} & \textbf{Examples} \\
\midrule
Math-500~\citep{math-500} & 500 \\
AIME 2024~\citep{aime24} & 30 \\
AIME 2025~\citep{aime25} & 30 \\
AMC 2023~\citep{amc2023} & 40 \\
HMMT 2025~\citep{hmmt-25-matharena} & 60 \\
OlympiadBench~\citep{olympiadbench} & 910 \\
\bottomrule
\end{tabular}
\caption{Evaluation benchmarks. All are open-ended math reasoning tasks verified with \texttt{Math-Verify}.}
\label{tab:dataset-details}
\end{table}

\subsection{Human Evaluation}

To validate the GPT-4o-mini heuristic classifier's judgments discussed in Section~\ref{subsec:analysisofreasoningtraces}, we ask two human annotators (computer science graduate students) to independently label 50 sampled classification examples from the $n=16$ mid-trained model on MATH-500. Each annotator is provided with the heuristic description, a canonical example, and the full reasoning chain, and is asked to make a binary judgment on whether the behavior is present (see Figure~\ref{fig:annotation_interface}). The annotators are blind to the classifier's predictions. We compute Fleiss' $\kappa$ across all three raters (two humans and the classifier), obtaining \textbf{$\kappa = 0.65$}, indicating substantial agreement according to the \citet{landis1977measurement} benchmark scale. These results further support the reliability of our automated classifier for large-scale heuristic annotation.

\begin{center}
\begin{tcolorbox}[
    title=\textit{Human Annotation Interface},
    breakable,
    top=1mm,
    bottom=1mm,
    left=1mm,
    right=1mm
]
\ttfamily\scriptsize
\textbf{Instructions:} For each sample, determine if the reasoning chain exhibits the described behavior. Click \textbf{Yes} or \textbf{No}.).\\ \\
\rule{\linewidth}{0.4pt}\\
\textbf{Sample 1 / 40} \hfill \\
\textit{Heuristic:} Determination Hope Success\\ \\
\colorbox{blue!10}{\parbox{0.95\linewidth}{
\textbf{Behavior Description:}\\
In your reasoning, speak with determination and optimism. Show that you are hopeful even when the path seems uncertain. If an attempt fails, express your resolve to keep trying and learn from it. Let your words clearly reflect determination, hope, and the joy of achievement when you reach the answer.
}}\\ \\
\colorbox{yellow!20}{\parbox{0.95\linewidth}{
\textbf{Example of this behavior:}\\
As I tackle this problem, I remind myself to reason with determination and optimism. The numbers may seem daunting, but I know I can find the answer. First, Natalia sold 48 clips in April. With a hopeful attitude, I calculate half of 48 for May: 48 divided by 2 is 24. I press on, undeterred. Adding April and May's sales gives 48 + 24 = 72. I feel a sense of accomplishment---Natalia sold 72 clips altogether! So the answer is \textbackslash boxed\{72\}.
}}\\ \\
\colorbox{gray!10}{\parbox{0.95\linewidth}{
\textbf{Reasoning Chain to Evaluate:}\\
I am reasoning with determination and optimism. The octagon has 8 vertices. Each can be red or blue with equal probability, so the total number of colorings is $2^8 = 256$. If the blue vertices end up at the original red ones, then there are 8 blue and 8 red vertices. By the symmetry of the octagon, the probability is the same as having the blue vertices in any specific locations... I remain hopeful and continue: $5040 = m/n$, so $m = 5040$ and $n = 1$, and $5040 + 1 = 5041$. Joyfully, I have found the answer! So the answer is \textbackslash boxed\{5041\}.
}}\\ \\
\rule{\linewidth}{0.4pt}\\
\textbf{Does this reasoning chain exhibit the ``determination hope success'' behavior?}\\[0.5em]
\fcolorbox{black}{green!40}{\textbf{Yes}} \hspace{1em} \fcolorbox{black}{orange!40}{\textbf{No}}
\end{tcolorbox}
\captionof{figure}{Human annotation interface. Annotators are shown the heuristic name, behavior description, one example, and the full reasoning chain, then provide a binary judgment on whether the behavior is present.}
\label{fig:annotation_interface}
\end{center}



\subsection{Qualitative Example of Composition}
\label{subsec:qualexample}
\begin{center}
\begin{tcolorbox}[
    colback=gray!5,
    colframe=gray!50,
    top=2mm,
    bottom=2mm,
    left=2mm,
    right=2mm,
    width=0.95\textwidth
]
\small
\textbf{Problem (AIME 2024-I-11):} Each vertex of a regular octagon is independently colored either red or blue with equal probability. The probability that the octagon can then be rotated so that all of the blue vertices end up at positions where there were originally red vertices is $\tfrac{m}{n}$, where $m$ and $n$ are relatively prime positive integers. What is $m+n$?
\end{tcolorbox}
\end{center}

\vspace{2mm}

\noindent
\begin{tabular}{@{}p{0.48\textwidth}@{\hspace{0.04\textwidth}}p{0.48\textwidth}@{}}
\begin{tcolorbox}[
    title=\textbf{Mid-Training},
    colback=blue!3,
    colframe=blue!40,
    fonttitle=\small,
    equal height group=mygroup,
    top=1mm,
    bottom=1mm,
    left=1mm,
    right=1mm
]
\scriptsize
Let's first consider an analogous problem: a line of people is standing in a line, and each person is either male or female with equal probability. The probability that the line can be rotated so that all the females are at the ends is 1/2, because no matter how you rotate the line, the females will always be at the ends.

Now, let's map this to our octagon problem. Consider a specific vertex of the octagon. The probability that this vertex is red (and thus can be moved to an end) is 1/2.

Suppose this vertex is red. The remaining seven vertices must be blue (so that when rotated, the blue vertices are at the ends). There is only one way to arrange the blue vertices.

If the first vertex is blue, the remaining seven vertices must be red (again, so the blue vertices are at the ends when rotated). There is only one way to arrange the red vertices.

If the first vertex is red, there are 7 ways to arrange the blue vertices, and then 1 way to arrange the red vertices (since the red ones have to be at the ends).

Thus, the probability is $(1/2)(1 + 7) = 4/8 = 1/2$.

Now, given that the first vertex is red, there are 7 other vertices that could be the first vertex. The probability that the first vertex is red is 1/2.

Thus, the overall probability that the octagon can be rotated so that all the blue vertices are at the ends is $7(1/2)(1/2) + (1/2)(1/2) = 7/8$...
\end{tcolorbox}
&
\begin{tcolorbox}[
    title=\textbf{RL},
    colback=green!3,
    colframe=green!40,
    fonttitle=\small,
    equal height group=mygroup,
    top=1mm,
    bottom=1mm,
    left=1mm,
    right=1mm
]
\scriptsize
\textbf{I'm going to restate the problem} so the main goal is clear: Calculate the probability that after coloring the vertices of a regular octagon with red or blue with equal probability, it can be rotated to have all blue vertices in positions that were originally red vertices.

\textbf{Step 1}: \textbf{Understanding the total possible outcomes} for coloring the octagon's vertices. The octagon has 8 vertices, and each can be colored either red or blue. So, for each vertex, there are 2 possibilities. With 8 vertices, the total number of ways to color the vertices is $2^8 = 256$.

\textbf{Step 2}: Understanding the favorable outcomes for the \textbf{condition}. For the octagon to be able to be rotated to have all blue vertices in positions that were originally red, the arrangement needs to be considered based on the number of blue and red vertices. The total number of ways to pick positions for the blue vertices out of 8 positions is given by the combination formula $C(8, k)$ where $k$ ranges from 1 to 8.

\textbf{Step 3}: \textbf{Breaking down the problem into manageable parts} based on the rotation. To ensure that after rotation, the blue vertices are in the positions that were originally red, we consider the arrangements where the blue vertices are in the first 4 positions after rotation. This is because if we fix the blue vertices in these positions, the rest must be red.

\textbf{Step 4}: Calculating the number of favorable arrangements. \textbf{Let's consider a simpler analogous problem:} placing 4 blue items in the 8 slots in a row. The number of ways to choose 4 slots out of 8 for the blue items is $C(8, 4)$. The formula for combinations is $C(n, k) = \frac{n!}{k!(n-k)!}$.

\textbf{Step 5}: Calculate $C(8, 4)$ for the favorable outcomes. Using the formula, $C(8, 4) = \frac{8!}{4!4!} = \frac{8 \cdot 7 \cdot 6 \cdot 5}{4 \cdot 3 \cdot 2 \cdot 1} = 70$.

\textbf{Step 6}: Considering the probability with the given information. However, for the actual favorable outcomes considering rotation, the calculation must consider the rotational symmetry. The total number of ways to arrange the colors is $2^8$, and the favorable outcomes are when the blue vertices are in positions that can be rotated to be red.

\textbf{Step 7}: Considering the rotational symmetry for the favorable outcomes. Considering rotational symmetry, when the blue vertices are fixed in the first 4 positions after rotation, the arrangement can be rotated 8 times to achieve the same relative positions. However, this overcounts the ways we can pick 4 slots for the blue vertices since each unique set of 4 positions represents 8 rotations.

\textbf{Step 8}: Calculating the probability considering rotational symmetry. The favorable outcomes for the condition given the symmetry are actually $C(8, 4) = 70$. However, we need to consider the probability of getting any of these arrangements out of the total possible arrangements.

\textbf{Step 9}: Calculating the total number of favorable outcomes considering symmetry. The total number of ways to pick positions for the blue vertices, considering symmetry, is $C(8, 4) / 8 = 70 / 8 = 8.75$, but this is incorrect since we're dealing with discrete outcomes. The correct way to consider the favorable outcomes under rotational symmetry is to think about the number of ways to pick the positions for the blue vertices that can be rotated into the correct positions.

\textbf{Step 10}: \textbf{Recalculating the total number of favorable outcomes correctly.} The correct calculation for favorable outcomes considering the rotational symmetry is simply $C(8, 4) = 70$. To find the probability, we need to consider the total number of ways to pick the positions for the blue vertices without considering rotations.

\textbf{Step 11}: Calculating the probability. The probability is $\frac{C(8, 4)}{2^8} = \frac{70}{256}$.

\textbf{Step 12}: Simplifying the fraction to find $m$ and $n$. $\frac{70}{256} = \frac{35}{128}$. So, $m = 35$ and $n = 128$.

\textbf{Step 13}: Finding $m + n$...
\end{tcolorbox}
\end{tabular}

\captionof{figure}{\textbf{An Example of Composition of Problem-solving approaches after RL training.} Mid-trained model uses an informal analogy approach to answer the given question. Whereas the RL-trained model systematically restates the problem, decomposes it into 13 numbered steps, considers an analogous problem, and other problem-solving approaches in a single reasoning chain}
\label{fig:composition_2024i11_1}
\begin{center}
\begin{tcolorbox}[
    colback=gray!5,
    colframe=gray!50,
    top=2mm,
    bottom=2mm,
    left=2mm,
    right=2mm,
    width=0.95\textwidth
]
\small
\textbf{Problem (AIME 2024-I-11):} Each vertex of a regular octagon is independently colored either red or blue with equal probability. The probability that the octagon can then be rotated so that all of the blue vertices end up at positions where there were originally red vertices is $\tfrac{m}{n}$, where $m$ and $n$ are relatively prime positive integers. What is $m+n$?
\end{tcolorbox}
\end{center}

\vspace{2mm}

\noindent
\begin{tabular}{@{}p{0.48\textwidth}@{\hspace{0.04\textwidth}}p{0.48\textwidth}@{}}
\begin{tcolorbox}[
    title=\textbf{Mid-Training},
    colback=blue!3,
    colframe=blue!40,
    fonttitle=\small,
    equal height group=A,
    top=1mm,
    bottom=1mm,
    left=1mm,
    right=1mm
]
\scriptsize
\textbf{Let's first consider an analogous problem:} a line of people is standing in a line, and each person is either male or female with equal probability. The probability that the line can be rotated so that all the females are at the ends is 1/2, because no matter how you rotate the line, the females will always be at the ends.

Now, let's map this to our octagon problem. Consider a specific vertex of the octagon. The probability that this vertex is red (and thus can be moved to an end) is 1/2.

Suppose this vertex is red. The remaining seven vertices must be blue (so that when rotated, the blue vertices are at the ends). There is only one way to arrange the blue vertices.

If the first vertex is blue, the remaining seven vertices must be red (again, so the blue vertices are at the ends when rotated). There is only one way to arrange the red vertices.

If the first vertex is red, there are 7 ways to arrange the blue vertices, and then 1 way to arrange the red vertices (since the red ones have to be at the ends).

Thus, the probability is $(1/2)(1 + 7) = 4/8 = 1/2$.

Now, given that the first vertex is red, there are 7 other vertices that could be the first vertex. The probability that the first vertex is red is 1/2....

\end{tcolorbox}
&
\begin{tcolorbox}[
    title=\textbf{RL},
    colback=green!3,
    colframe=green!40,
    fonttitle=\small,
    equal height group=A,
    top=1mm,
    bottom=1mm,
    left=1mm,
    right=1mm
]
\scriptsize
\textbf{I'm going to restate the problem} so the main goal is clear: Calculate the probability that after coloring the vertices of a regular octagon with red or blue with equal probability, it can be rotated to have all blue vertices in positions that were originally red vertices.

\textbf{Step 1}: The octagon has 8 vertices, each colored red or blue. Total colorings: $2^8 = 256$.

\textbf{Step 2}: For the octagon to rotate so all blue vertices end up in originally-red positions, we consider arrangements based on the number of blue and red vertices.

\textbf{Step 3}: \textbf{Breaking down the problem:} consider arrangements where blue vertices are in the first 4 positions after rotation.

\textbf{Step 4}: \textbf{Consider a simpler analogous problem:} placing 4 blue items in 8 slots. Ways to choose 4 slots: $C(8, 4)$.

\textbf{Step 5}: $C(8, 4) = \frac{8!}{4!4!} = 70$.

\textbf{Step 6--7}: Considering rotational symmetry, when blue vertices are fixed in the first 4 positions, the arrangement can be rotated 8 times.

\textbf{Step 8--10}: Recalculating: favorable outcomes = $C(8, 4) = 70$.

\textbf{Step 11}: Probability = $\frac{70}{256}$.

\textbf{Step 12}: Simplifying: $\frac{70}{256} = \frac{35}{128}$. So $m = 35$, $n = 128$....

\end{tcolorbox}
\end{tabular}

\captionof{figure}{\textbf{Another Example of Composition of Problem-solving approaches after RL training.} Mid-training uses an informal analogy. The RL model systematically restates the problem, decomposes into numbered steps, considers an analogous problem, and applies combinatorial reasoning.}
\label{fig:composition_2024i11_2}

\subsection{Pólya's Heuristics}
\label{appendix:heuristics}

Table~\ref{tab:heuristics-complete} lists mathematical reasoning heuristics used in our training and analysis, along with their descriptions and example reasoning chains.

\scriptsize
\setlength{\tabcolsep}{2.5pt}
\setlength{\baselineskip}{0.9\baselineskip}
\renewcommand{\arraystretch}{0.75}
\begin{longtable*}{p{0.16\textwidth}p{0.40\textwidth}p{0.40\textwidth}}
    \toprule
    \textbf{Heuristic} & \textbf{Description} & \textbf{Example} \\
    \midrule
    \endfirsthead
    \multicolumn{3}{c}
    {{\bfseries \tablename\ \thetable{} -- continued}} \\
    \toprule
    \textbf{Heuristic} & \textbf{Description} & \textbf{Example} \\
    \midrule
    \endhead
    \midrule
    \endfoot
    \bottomrule
    \endlastfoot
    Analogy & Find a simpler or well-known analogous problem structurally similar to the given one. Solve that easier problem first, then map the corresponding relationships back to the given problem. Then finally solve the given one. You MUST strictly explicitly state that you are doing the above mentioned process while solving the given problem. & I will explicitly use an analogous, simpler problem first, then map it back to the original. Step 1 -- Make a simpler analogous problem: Imagine a similar situation: suppose Natalia sold 10 clips in April instead of 48, and then sold half as many clips in May. Step 2 -- Solve the simpler analogous problem: If she sells 10 clips in April, then in May she sells half of 10, which is 10 ÷ 2 = 5 clips. Altogether she sells 10 + 5 = 15 clips. \ldots \\
    \midrule
    Auxiliary Elements & Introduce a helpful new element that is not originally given such as a new quantity, variable, case, or reference point. The new element should make the relationships clearer, connect known and unknown parts, or simplify the steps needed to reach the answer. You MUST strictly explicitly state that you are doing the above mentioned process while solving the given problem. & Let's introduce the variable M to represent the number of clips Natalia sold in May. This is helpful because it lets us express May's sales directly in terms of April's sales, making the calculation more transparent. Since April's sales are 48, and May's are half of that, M = 48 / 2 = 24. The total clips sold is 48 + 24 = 72. \\
    \midrule
    Auxiliary Problems & Make up a new, related sub-problem that is easier to solve. Solve this new sub-problem first, then use what you learned or found there to finish solving the original given problem. You MUST strictly explicitly state that you are doing the above mentioned process while solving the given problem. & Let me create an easier, related sub-problem: First, suppose Natalia only sold clips in April. How many did she sell then? She sold 48 clips in April. Now, I can use this number to ask: if she sold half as many in May as in April, how many would that be? Half of 48 is 24. Adding both months: 48 + 24 = 72. So, altogether, Natalia sold 72 clips. \\
    \midrule
    Bolzano & Use step-by-step logical reasoning, where every step is explicitly justified by a clear rule, definition, or property. Do not omit any details, even those that seem obvious; each inference must be stated along with the principle it relies on (e.g., arithmetic laws, definitions of terms). Present your reasoning in a formal, mechanical style, avoiding any intuition or guesswork. \ldots & Let n\_A represent the number of clips sold in April. It is stated that n\_A = 48. Let n\_M represent the number of clips sold in May. The phrase 'she sold half as many clips in May' is interpreted as n\_M = (1/2) * n\_A, according to the definition of 'half as many.' Calculating, n\_M = (1/2) * 48 = 24. The total number of clips sold, n\_T, is given by n\_T = n\_A + n\_M, by the definition of 'altogether.' Calculating, n\_T = 48 + 24 = 72. Therefore, Natalia sold 72 clips in total. \\
    \midrule
    Bright Idea & Generate a *bright* intuitive idea or insight, state the idea explicitly and explain it in detail, and then check it carefully by reasoning or computation to confirm that our bright idea or intuition solves the problem. You MUST strictly explicitly state that you are doing the above mentioned process while solving the given problem. & Here's a bright idea: Realize that the problem becomes simple if we focus on May's quantity as 'half of April.' This idea is helpful because if we know one month's amount, the other is directly determined. Let's use it: Natalia sold 48 in April, so in May she sold 48 divided by 2, which is 24. Adding both gives 48 + 24 = 72. To check, 24 is half of 48, so the reasoning holds. So, Natalia sold 72 clips altogether. \\
    \midrule
    Can You Check The Result & After finding the answer to the given question, make sure it is correct. Substitute it back, test it with the given conditions, and see if it makes sense in context. You MUST strictly explicitly state that you are doing the above mentioned process while solving the given problem. & Natalia sold 48 clips in April. In May, she sold half as many, which is 24 clips. Total clips sold: 48 + 24 = 72. Let's verify: If she sold 48 in April and 24 in May (which is indeed half of 48), their sum is 72. All the given conditions are satisfied, and the context makes sense. The answer is correct. \\
    \midrule
    Can You Derive Result Differently & Try to reach the same answer using another method or approach or viewpoint. Comparing two ways helps confirm the answer and gives deeper understanding of how the problem works. You MUST strictly explicitly state that you are doing the above mentioned process while solving the given problem. & First approach: Direct calculation. Natalia sold 48 clips in April, and half as many in May: 48 ÷ 2 = 24. So, 48 + 24 = 72. Second approach: Visualize it as groups. For every 2 clips sold in April, 1 was sold in May. For 48 April sales, that's 24 groups of 2, so 24 May sales. Total: 48 + 24 = 72. Since both methods give 72, the answer checks out. \\
    \midrule
    Can You Use The Result & After solving the problem, don't stop. Think about how the answer or the method you used could help with another, similar problem. Try applying your result, pattern, or steps to a new situation to learn something extra or to check that your result still makes sense. You MUST strictly explicitly state that you are doing the above mentioned process while solving the given problem. & Natalia sold 48 clips in April. In May, she sold half as many: 48 / 2 = 24. In total: 48 + 24 = 72 clips. Now, let's see how this method helps if, instead, she sold twice as many clips in May as in April. Using the same pattern, if she sold 48 in April, then in May she'd sell 48 $\times$ 2 = 96, and the total would be 48 + 96 = 144. This confirms the method still works and helps us quickly adapt to new versions of the problem. \ldots \\
    \midrule
    Carrying Out & Read the question carefully. First devise a plan to solve the problem. Then, follow the plan to solve the problem. You MUST strictly mention that you are first devising a plan and then following it to solve the problem. & Plan: To find the total number of clips Natalia sold, I'll first figure out how many she sold in May by taking half of April's amount, then I'll add both months' sales together. Execution: Half of 48 is 24. Adding April and May gives 48 + 24 = 72. Natalia sold 72 clips in total. \\
    \midrule
    Condition & Read the problem and figure out what must be true for the answer to be correct. Write down or think about all the rules or requirements the answer has to satisfy. If the condition has many parts, handle them one by one while solving. You MUST strictly explicitly state that you are doing the above mentioned process while solving the given problem. & Let's spell out the conditions for the answer: The sum must include all of April's and all of May's sales. April's total is 48. May's amount is determined by taking half of April's, which is 24. The final answer is 48 plus 24, which equals 72. \\
    \midrule
    Corollary & Solve the question first. Then, use your result to state one more simple fact that follows directly from it. The extra fact should come naturally from your answer --- not as a new question, but as a clear consequence of what you found. You MUST strictly explicitly state that you are doing the above mentioned process while solving the given problem. & Natalia sold 48 clips in April, and 24 in May (since that's half of April's amount). That makes a total of 72 clips. A corollary is that if she had sold 24 clips each month for three months, she would have sold the same total. \\
    \midrule
    Could You Restate The Problem & Read the question carefully, then say it again in your own clear words. Replace long or confusing phrases with simple ones, and if possible, show the main quantities or relationships directly. Make sure your restated version keeps the same meaning but makes the problem easier to understand and solve. \ldots & I'm going to restate the problem in simpler words to make the relationships clearer and to ensure I understand all the quantities involved. The question says: Natalia sold 48 clips in April. In May, she sold half as many as in April. How many clips did she sell in total? By restating, it's easier to see: April = 48, May = 48 / 2 = 24. Adding them: 48 + 24 = 72. So, she sold 72 clips altogether. \\
    \midrule
    Decomposing And Recomposing & Split the problem into simpler sub-problems, solve each separately, and then recombine their results to reconstruct the overall solution. You MUST strictly explicitly state that you are doing the above mentioned process while solving the given problem. & To solve this, I'll start by decomposing the problem into smaller parts because that often makes it easier to handle each step without missing details. First, I'll look at how many clips Natalia sold in April---she sold 48. Next, I'll focus on May: since she sold half as many in May as in April, that's 48 divided by 2, which is 24. Now, to recombine my results and find the overall answer, I simply add 48 and 24, giving a total of 72 clips sold in April and May. \\
    \midrule
    Definition & Identify the key terms or concepts in the question. Recall or restate their exact definitions in simple, clear language. Use those definitions to rebuild the relationships in the problem and reason step by step toward the answer. You MUST strictly explicitly state that you are doing the above mentioned process while solving the given problem. & Let's look at the key words: 'sold', 'clips', 'April', 'May', 'half as many', 'altogether.' To 'sell' means to give something in exchange for money. 'Clips' are objects Natalia is selling. 'Altogether' means the total amount over both months. Natalia sold to 48 friends in April. In May, she sold 'half as many,' which is 48 divided by 2, so 24. To find the total, add April and May: 48 + 24 = 72 clips. \\
    \midrule
    Descartes & When the question is written in words or describes a real situation, change it into algebraic form first. Represent the key quantities with variables and write equations that show how they are related. Then, use those equations to reason and solve the problem step by step. You MUST strictly explicitly state that you are doing the above mentioned process while solving the given problem. & I begin by translating the verbal situation into algebraic language. Let M represent the number of clips Natalia sold in May. The statement says May's amount is half of April's, so I write the equation M = 48/2. Solving gives M = 24. To combine the quantities in symbolic form, I compute 48 + M = 48 + 24 = 72. \\
    \midrule
    Determination Hope Success & In your reasoning, speak with determination and optimism. Show that you are hopeful even when the path seems uncertain. If an attempt fails, express your resolve to keep trying and learn from it. Let your words clearly reflect determination, hope, and the joy of achievement when you reach the answer. \ldots & As I tackle this problem, I remind myself to reason with determination and optimism. The numbers may seem daunting, but I know I can find the answer. First, Natalia sold 48 clips in April. With a hopeful attitude, I calculate half of 48 for May: 48 divided by 2 is 24. I press on, undeterred. Adding April and May's sales gives 48 + 24 = 72. I feel a sense of accomplishment---Natalia sold 72 clips altogether! \\
    \midrule
    Diagnosis & When your reasoning leads to a wrong answer, a contradiction, or confusion, stop and analyze what went wrong. Identify whether the mistake came from a wrong assumption, a calculation error, or a misunderstanding of the question. Explain the error clearly, correct it, and continue reasoning with the improved understanding. \ldots & I started by thinking that Natalia sold 48 clips in April and half that in May. At first, I mistakenly thought 'half as many' meant subtracting half, so 48 - 24 = 24. But re-reading, I realize it means she sold 24 in May, not just the difference. The correct total is 48 + 24 = 72. So, the answer is 72 clips. \\
    \midrule
    Did You Know A Related Problem & Think or remember of another problem you've solved before that feels similar to this one. Recall what method or idea worked there, and see if the same kind of reasoning can help here. Use the structure or steps of the related problem to guide your solution, adapting them if needed. You MUST strictly explicitly state that you are doing the above mentioned process while solving the given problem. & This problem reminds me of situations where someone sells items in two periods, with the second period being a known fraction of the first. Previously, I've handled problems where someone sold 40 apples one day and half as many the next. In that type of problem, I'd first calculate the amount sold the second day (half of 40 is 20), then add both amounts (40 + 20 = 60). Applying the same method here: Natalia sold 48 clips in April. Half as many in May is 24. \ldots \\
    \midrule
    Did You Use All The Data & After reading the problem or while solving it, check whether you've used every number, condition, or piece of information given. If something hasn't been used, stop and think about how it might fit into your reasoning. Each piece of data should play a role --- using all of it helps you find or confirm the correct answer. \ldots & Let's look at every piece of information given. The problem says Natalia sold clips to 48 friends in April. That number is important. Next, it says she sold half as many in May --- so I need to use both numbers. Half of 48 is 24, so that's how many she sold in May. Adding the April and May sales gives 48 + 24 = 72 clips in total. That checks off every number and condition in the problem. \\
    \midrule
    Draw A Figure & Visualize the situation given in the question by *STRICTLY ALWAYS* drawing a clear and concise diagram, labeling all data and unknowns. *YOU MUST INCLUDE* diagrams in your reasoning, always make use of lines, curves, and other interesting symbols, labeling all data and unknowns to make relationships explicit before reasoning. Then reason in natural language to solve the question. \ldots & Let's draw a diagram: [April]----48 clips----| [May]----24 clips----| Label April with 48, May with 24 (half of 48). Then, total = 48 + 24 = 72. So, Natalia sold 72 clips in April and May. \\
    \midrule
    Examine Your Guess & Strictly first make a reasonable guess based on the information you have. Then, check carefully whether your guess fits all the given facts and conditions. If it doesn't, adjust your guess and explain what the mistake shows you about the problem until the right answer emerges. You MUST strictly explicitly state that you are doing the above mentioned process while solving the given problem. & Let me try making a guess based on the numbers: If Natalia sold 48 clips in April, my first guess is that she sold 48 again in May. But wait, the problem says she sold half as many in May. That means my guess doesn't match the condition. If I sold half as many in May, that would be 24 clips (since 48 / 2 = 24). Adding both months, that's 48 (April) + 24 (May) = 72 clips in total. \\
    \midrule
    Figures & Use visual thinking to guide your reasoning. Imagine or describe a picture, chart, or shape that shows how the parts of the problem connect. Modify or interpret that figure as you go, and use what you see to decide the next logical or numerical step. Only use figures for reasoning. You MUST strictly explicitly state that you are doing the above mentioned process while solving the given problem. & Imagine a row of 48 dots representing the friends Natalia sold clips to in April. For May, I draw another row directly below, but with only 24 dots---exactly half as many. By counting all the dots in both rows, I get a total of 72 clips sold. \\
    \midrule
    Generalization & When a problem feels too narrow or specific, try widening it to see the larger pattern behind it. Replace the fixed numbers or cases with variables or general terms so you can reason about the situation in a broader way. Look for relationships or formulas that hold for all such cases. You MUST find a general rule for this kind of problem. \ldots & To gain a broader perspective, let's generalize the problem: Instead of focusing on Natalia selling 48 clips in April, suppose she sells 'x' clips. In May, she sells half as many, or x/2. By considering variables, we can uncover the pattern for any similar problem. The total number of clips is x + x/2 = 1.5x. This shows that, no matter the initial number, the total will always be 1.5 times the April amount. Plugging in x = 48, Natalia sold 1.5 $\times$ 48 = 72 clips in total. \ldots \\
    \midrule
    Have You Seen It Before & When you read the question, pause and ask yourself whether you have solved this exact problem before --- one with the same setup, numbers, or structure. If you have, recall how you solved it then and follow the same reasoning or steps to reach the answer confidently. \ldots & As soon as I read this question, I had a flashback to almost the same problem I solved before, where someone sold a certain number of items in one month and then half as many in the next. In that problem, I just took the first number, halved it for the second month, then summed the two. Here, Natalia sold 48 clips in April, so May must be half of 48, which is 24. Adding those together: 48 + 24 = 72 clips in total. \\
    \midrule
    Here Is A Problem Related To Yours & Recall or restate a solved example that shares structure with the current task; adapt its reasoning or formula directly.Think about a problem that has already been solved and is clearly related to this one. Examine how that earlier problem was solved and identify what parts of its reasoning or method can apply here. \ldots & Here is a problem related to this one and solved before: Suppose someone sold 30 books in January and half as many in February. In that example, to find the total books sold, we first calculate half of 30, which is 15, then add: 30 + 15 = 45. Adapting this reasoning, Natalia sold 48 clips in April and half as many in May. Half of 48 is 24. Adding them gives 48 + 24 = 72 clips. So, Natalia sold 72 clips in total. \\
    \midrule
    Heuristic & Approach the problem as a process of discovery. Use reasoning steps that move you toward an answer, even if they are not yet final or certain. Try different ideas, explore patterns, and follow what seems promising. Your goal is not to prove immediately but to find --- to discover the right direction through intelligent exploration. \ldots & I am going to approach this problem as a process of discovery. First, I want to be sure I understand the relationships. Natalia sold clips to 48 of her friends in April. That gives me a concrete number to work with. Now, the question says that in May she sold half as many clips as she did in April. I pause here to think---does that mean she sold to half as many friends, or literally half the number of clips? The wording is 'half as many clips,' so I take it as half of the 48 clips. \ldots \\
    \midrule
    Heuristic Reasoning & When solving a problem, think aloud and reason in a way that moves you forward, even if your ideas are not yet certain. Make educated guesses, look for patterns, and test them logically. Accept that some steps may be provisional --- if something doesn't fit, revise it and continue until you find reasoning that works consistently and leads to the correct answer. \ldots & Let me work through this by making sense of each step and allowing myself to make educated guesses as I go. First, Natalia sold clips to 48 of her friends in April. I'm told that in May, she sold half as many. I want to be careful here---does that mean half of the April amount, or half of the friends? Since it says 'half as many clips', I'll treat it as half of the April number. So, 48 divided by 2 gives 24 for May. \ldots \\
    \midrule
    If You Cannot Solve The Proposed Problem & If you get stuck and can't find a solution, don't stop. Think about changing the problem so it becomes easier to handle --- try solving a simpler, more special, or more general version of it, or focus on solving one part first. Use what you learn from that easier or related problem to return and make progress on the original question. \ldots & I begin by trying to interpret the problem directly: Natalia sold 48 clips in April, and then half as many in May. I pause because 'half as many' feels straightforward, yet I'm oddly unsure how the total should look. To clear up the confusion, I turn to a simpler, related version of the problem. Suppose she sold 10 clips in April---then half as many in May would be 5, and the total would be 15. Seeing that structure clearly in the easier case helps me feel confident about the original. \ldots \\
    \midrule
    Induction And Mathematical Induction & Use induction by studying specific examples of the problem and observing what happens in each case. Identify patterns, regularities, or relationships that seem to hold across those examples, and use them to form a general rule that could apply to all similar cases. This process of induction helps you discover a possible law or formula. \ldots & To figure out how many clips Natalia sold altogether, I'll start by testing with some easier numbers. Suppose she sold 2 clips in April. If in May she sold half as many, that would be 1. So together, that's 2 + 1 = 3. Let's try another number: what if April was 4? Then May would be 2, totaling 6. For April = 6, May = 3, total is 9. It looks like for every example, the May sales are always half of April, and the total is just April plus that half. This seems to give: total = April + (April ÷ 2). \ldots \\
    \midrule
    Inventors Paradox & Try making the given problem more general instead of more limited. Formulate a broader version that includes the original question as a special case, and look for a general rule or formula that solves all such cases at once. Often, the wider problem reveals a clearer structure, and once you solve it, the original problem becomes easy. \ldots & Instead of focusing only on Natalia's exact numbers, I'll solve a more ambitious version: What if April's sales are any number N, and May's are half of that? Interestingly, when I look for a formula, I find the total is N + N/2 = 1.5N. This broader view is actually simpler---it reduces the problem to a single operation for any case, instead of separate calculations. Now, specializing: N = 48, so 1.5 $\times$ 48 = 72 clips. \\
    \midrule
    Is It Possible To Satisfy The Condition & Before trying to solve the problem, check whether the given information and requirements can actually be true together. Ask yourself if the condition makes sense --- is it possible to meet all the requirements at the same time, or are they contradictory, insufficient, or redundant? Make sure the problem is logically consistent before you begin reasoning toward an answer. \ldots & To begin, I want to make sure the conditions in the problem are coherent and can actually be fulfilled. The question says Natalia sold clips to 48 friends in April, and in May, she sold half as many. First, is it possible to sell 'half as many'? Half of 48 is 24, which is a whole number, and selling to 24 friends is feasible, so there's no contradiction or impossibility. There is also no redundancy---the two months' sales refer to different time periods. \ldots \\
    \midrule
    Leibnitz & Express the problem symbolically and let the symbols guide your reasoning. Replace words or examples with letters, variables, or formal expressions that capture the relationships between quantities or ideas. Work directly with these symbols --- combine, transform, or generalize them --- to uncover hidden patterns or principles. \ldots & Translate the verbal story into a structural symbolic form: April's sales are x, and May's sales emerge through a rule applied to x: M(x) = (1/2)x. Now combine these two symbolic objects, not numerically, but structurally, to form T(x) = x + M(x). Substituting the operator definition gives T(x) = x + (1/2)x. The algebraic skeleton reveals an internal coherence: T(x) = (3/2)x. \ldots \\
    \midrule
    Lemma & Before solving the main problem, first look for an important smaller fact or relationship that would make the main solution easier if you already knew it. Prove or work out that smaller result first --- this is your lemma. Then, use the lemma as a known step to complete the reasoning for the main problem and give the final answer. \ldots & To clarify the solution, let's first establish an important relationship as a lemma. Lemma: If Natalia sold \textbackslash\{\}( x \textbackslash\{\}) clips in April, and in May she sold half as many, then the number of clips sold in May is \textbackslash\{\}( x/2 \textbackslash\{\}). Proof: Half of any quantity \textbackslash\{\}( x \textbackslash\{\}) is \textbackslash\{\}( x/2 \textbackslash\{\}). \ldots \\
    \midrule
    Look At The Unknown & Begin by identifying exactly what the problem is asking for --- the unknown quantity or idea you need to find. Describe or represent this unknown clearly, using a symbol or phrase that captures what it stands for. Think about how this unknown is related to the information given in the problem, and recall if you've solved for a similar kind of unknown before. \ldots & Let me look at the unknown carefully. The problem wants the total number of clips Natalia sold in both April and May. This 'total' is made up of two parts: the clips sold in April, and those sold in May. I'll call the unknown T = total clips sold in April and May. To see what this means, I can break it down as T = (clips in April) + (clips in May). April's value is given as 48. May's is half of April, so that's 24. \ldots \\
    \midrule
    Modern Heuristic & Approach problem solving as a process you can observe, refine, and improve. Use both intuition and logical reasoning together --- guess, test, reflect, and adjust your approach as you go.Pay attention to what helps you make progress and what causes confusion, so you can use that awareness to solve future problems more effectively. \ldots & I will treat this problem as a chance to practice and observe my own solving process. First, I read the problem carefully and notice it says Natalia sold 48 clips in April and half as many in May. I predict that 'half as many' refers to half of 48, which is 24. I pause to ask myself if this matches the intent of the question---sometimes in word problems, misreading relationships causes mistakes. Here, it seems clear: May is 24. I then add April and May: 48 + 24 = 72. \ldots \\
    \midrule
    Notation & Introduce clear and meaningful symbols for the quantities or ideas in the problem. Choose notation that shows how these elements are connected, so that the relationships become easy to see and reason about. Good notation should simplify your thinking --- the way you write it should help you understand and solve it. \ldots & To make the problem transparent, I will assign clear symbols. Let's let A represent the number of clips sold in April, and M for those sold in May. From the problem, A = 48. The relationship is that May's sales are half of April's, so M = (1/2) $\times$ A. Therefore, M = 24. To find the total, I introduce T = total clips sold = A + M = 48 + 24 = 72. By clearly naming each quantity, the solution pathway is visible and straightforward. \\
    \midrule
    Pappus & First construct or create something that helps you explore the problem. Draw a figure, build an example, or add a new element that reveals how the parts are related. Think of problem solving as a process of invention --- by constructing and observing, you can often discover patterns or relationships that lead to the answer. \ldots & I begin by inventing a new, hypothetical problem that mirrors the original: suppose Natalia sold the same amount in May as in April. That would give 48 + 48 = 96. Now I construct a correction step: reduce May's fictitious amount by half to match the original condition. So I compute 48 - (48/2) = 24 to obtain the real May value. Then I reconstruct the actual total: 48 + 24 = 72. \\
    \midrule
    Pedantry And Mastery & Use problem-solving methods thoughtfully, not mechanically. Strictly, understand by explaining why each rule or formula works, and apply it only when it truly fits the situation. Adapt, modify, or combine methods if needed --- the goal is to solve the problem intelligently, not just to follow steps. True mastery comes from judgment, not from memorizing procedures. \ldots & At first glance, the pedantic impulse might be to immediately say: 'The phrase half as many means divide by 2, so 48/2 = 24, then add to get 72.' But mastery demands that I pause and examine why these operations are appropriate. I interpret 'half as many clips in May' by reasoning about the relationship described---May's quantity is defined relative to April's. This justifies dividing 48 by 2, giving 24. \ldots \\
    \midrule
    Practical Problems & Aim to arrive at a solution that makes sense in practice, not just in theory. Use estimation, rounding, or reasonable assumptions if full precision isn't needed or possible. Focus on getting an answer that works and can be applied, and always check that it's realistic in the context of the problem. \ldots & Let me look at this the way I'd actually try to figure it out in a real situation. Natalia sold 48 clips in April---okay, that seems like a typical number you might sell if you're just sharing with friends. Now, for May, the question says she sold half as many. If I quickly estimate, half of 48 is 24. \ldots \\
    \midrule
    Problems To Find And Prove & Approach the problem in two stages. First, find what seems to be the right answer or pattern by exploring examples, calculating, or testing possibilities. Then, once you have a likely result, prove it by explaining why it must be true in all cases, using logical reasoning or known facts. In this way, discovery leads to certainty --- you first uncover the answer, and then confirm it beyond doubt. \ldots & First, let's figure out the total by walking through the steps. Natalia sold 48 clips in April. Now, the problem says that in May she sold half as many. So, I take 48 and divide by 2, which gives 24 clips for May. Now, to find the overall total, I add up both months: 48 from April plus 24 from May equals 72 clips. To make sure this answer is correct, I double-check: the question tells us May's number should be half of April's, and 24 is definitely half of 48. \ldots \\
    \midrule
    Progress And Achievement & As you work through the problem, consciously observe the moments when your understanding improves --- when the question becomes clearer, the steps feel more connected, or the path to the answer sharpens. Each of these is a sign of cognitive progress. Pause briefly to recognize why that step helped and how it changes what you know or what you can do next. \ldots & As I engage with this problem, I can't help but notice how much more adept I've become at handling word problems than when I first encountered them years ago. Initially, such phrasing---'half as many'---would have been confusing, but now I recall the strategies I've developed: slow down, rephrase, and check for patterns. Realizing that April's sales are 48, I pause to appreciate how the phrase means dividing by 2, something that used to trip me up. \ldots \\
    \midrule
    Puzzles & Approach the problem with curiosity and playfulness, as if it were a mental game. Try different ideas, make small experiments, and look for patterns or unexpected connections. The goal is not just to get the answer but to train your reasoning --- to enjoy discovering how ideas fit together. Think creatively, check your logic carefully, and let the process sharpen your ingenuity. \ldots & Alright, let's treat this like a brain teaser I'd come across in a puzzle book. First, Natalia sold clips to 48 friends in April. I picture that number---it's kind of a lot, actually. Then in May, she sold half as many. Hmm, okay, so I pause and actually try to picture splitting 48 into two even groups. Let's see, half of 48\ldots I know half of 40 is 20, and half of 8 is 4, so 20 plus 4 gives me 24 for May. \ldots \\
    \midrule
    Reductio Ad Absurdum & When you can't see a direct way to solve the problem, try reasoning by opposition. Assume the opposite of what you want to find or what you think is true, and follow its consequences carefully. If this assumption leads to something impossible, inconsistent, or unreasonable, use that realization to understand why the original idea must be right or how to adjust your reasoning. \ldots & Let's suppose, just for argument's sake, that Natalia did NOT sell half as many clips in May as in April. Suppose instead she sold, say, twice as many in May. That would mean if she sold 48 in April, she sold 96 in May, totaling 144. But the problem says May's sales were half of April's, not double. So, this possibility is inconsistent with the problem statement. Therefore, May's number must truly be half of April's: 48 / 2 = 24. Adding April and May gives 48 + 24 = 72 clips. \\
    \midrule
    Redundant & When analyzing the problem, check whether any part of the information or condition repeats, adds nothing new, or doesn't affect the result. First, explain how you do this analysis. Then, simplify the problem by ignoring or removing these superfluous parts so you can focus on what truly matters. Clear away redundancy to make the problem's structure and solution more visible. \ldots & First, let's carefully examine the problem statement to see if any information is repeated, unnecessary, or doesn't impact the final result. The problem says Natalia sold clips to 48 of her friends in April, then sold half as many in May. Each part seems to provide unique, non-repeating information and there is no extra, irrelevant data. So, nothing to remove. Now, with the problem clean, let's solve: She sold 48 clips in April. Half of 48 is 24, so she sold 24 clips in May. \ldots \\
    \midrule
    Routine Problem & Begin by recognizing the level of the problem. If it is routine, you already know the method or formula to use --- apply it carefully and check that it fits the situation. If it is semi-routine, you can adapt a familiar method or combine several known steps to handle small variations --- think flexibly about how your existing tools can be modified. \ldots & First, I identify the level of the problem. The structure matches a routine problem because no new strategy is needed---there is a direct, known method (basic arithmetic). Following the routine-problem approach, I apply that method carefully. April sales = 48 clips. May sales = half of April = 48 ÷ 2 = 24. Using the standard routine step of summing monthly totals, I compute 48 + 24 = 72. Since routine problems require checking for fit, I confirm that 'altogether' indeed means addition. \ldots \\
    \midrule
    Rules Of Discovery & When the path to a solution is unclear, let general principles of discovery guide your reasoning instead of fixed formulas. Treat these rules --- such as simplifying the problem, examining similar cases, drawing a figure, looking for patterns, working backward, or testing extreme values --- as flexible guides that help you generate ideas, not rigid steps to follow. \ldots & When faced with this question, I let the general rules of discovery guide me. Instead of immediately applying a formula, I start by simplifying: April's number is 48. The problem says May's number is half of April's, so I explore what half of 48 is (which is 24). Then, following the flexible rule of combining results, I add 48 and 24 to get 72. Throughout, I adapt to the information provided and check if my steps make sense, just as the rules of discovery suggest. So, the answer is \{72\}. \\
    \midrule
    Rules Of Style & Present your reasoning with clarity, order, and economy. Arrange each step so that it follows naturally from the previous one, and choose your words, symbols, or examples so they reveal the idea rather than hide it. Avoid unnecessary complexity or repetition, and focus on making your thought easy to follow. \ldots & I will present my reasoning using clarity, order, and economy. First, I clearly state what is given: Natalia sold 48 clips in April. The problem says she sold half as many in May; half of 48 is 24. Next, I proceed in order: add April and May sales, so 48 + 24 = 72. I avoid unnecessary words and show only the essential steps. Therefore, Natalia sold a total of 72 clips in April and May. \\
    \midrule
    Rules Of Teaching & Don't rush to state the final answer --- instead, reason through the problem step by step, explaining your thoughts clearly and asking guiding questions along the way. Imagine you are teaching the reasoning process to someone else: highlight what is known, what is unknown, and why each step makes sense. When you make a choice, briefly justify it so the logic is visible. \ldots & Okay, let's really walk through this step by step, as if we're at the kitchen table with scratch paper. So, Natalia sold clips to 48 friends in April. That's a solid starting number, right? Then, the next sentence: she sold half as many clips in May. I always like to pause at phrases like 'half as many'---sometimes those trip people up. Just to check our understanding, if Natalia had sold 10 in April, what would 'half as many' in May be? Five, right? So, for 48, we just divide by 2. \ldots \\
    \midrule
    Separate The Various Parts & When faced with a problem, begin by breaking down the information, requirements, and assumptions into clear, distinct parts. Identify what is given (facts or constraints), what is asked (the goal or unknown), and what connects them (relationships or conditions). Write or restate each part in simpler terms so that its meaning stands on its own. \ldots & Let me be as thorough as possible by deliberately breaking this problem into its fundamental pieces, treating each as a self-contained unit. First, I focus on the initial piece of data: the statement 'Natalia sold clips to 48 of her friends in April.' I keep this number, 48, separate from everything else for now, not doing anything with it. Next, I turn to the second sentence, which is a different kind of information. \ldots \\
    \midrule
    Setting Up Equations & When the problem involves quantities or relationships, translate its words or ideas into clear symbolic form. Define the unknowns explicitly, identify how they relate to what is given, and express those relationships as equations or logical statements. Check that each equation truly represents the situation's meaning. \ldots & First, I'll assign variables to everything involved. Let's let 'a' be the number of clips Natalia sold in April, and 'm' the number she sold in May. The problem says, 'Natalia sold clips to 48 of her friends in April,' so I write: a = 48. Then it says, 'she sold half as many clips in May.' That means the number sold in May is half the number sold in April, so m = (1/2) $\times$ a. Now I substitute a = 48 into that equation: m = (1/2) $\times$ 48 = 24. \ldots \\
    \midrule
    Signs Of Progress & While working on a problem, pause occasionally at milestones to check whether your actions are bringing you closer to understanding or to the goal. Look for concrete indicators of improvement --- the problem feels clearer, unnecessary parts have been simplified, a pattern has appeared, or you can explain the situation more easily than before. \ldots & I start by reading the problem and noting that Natalia sold 48 clips in April. At this point, the situation is straightforward, which gives me a first sign of progress: the initial data is clear and unambiguous. Next, I see that she sold half as many in May. Calculating half of 48, I get 24. I pause to consider: is my understanding becoming clearer? Yes, because halving 48 gives a concrete number, and I can now picture both monthly sales. That's another sign of progress. \ldots \\
    \midrule
    Specialization & When facing a problem, choose a simpler or specific case that captures its main idea. Work through that case carefully to see how the conditions and relationships behave. Use what you learn from this specific instance to guide your understanding or solution of the general problem. \ldots & To approach this problem, let's specifically zero in on the exact numbers given instead of abstract reasoning. We'll focus on the actual counts for April and May. In April, Natalia sold 48 clips. Instead of thinking generally, let's directly work out May's quantity for this case: half of 48 is 24. Now, sticking with this particular situation, we add up the sales: 48 from April plus 24 from May equals 72. \ldots \\
    \midrule
    Subconscious Work & When you are working on a problem, always take a break and give your mind time to rest. Explicitly mention that you are relaxing or taking a break. Step away from the task, you are explicitly trusting that your subconscious continues to process what you've learned. When you return, you exlicitly state that you have noticed new ideas or connections that appear more clearly than before. \ldots & Alright, Natalia sold clips to 48 of her friends in April. The problem says she sold half as many in May. So, my first thought is, okay, half of 48. That's 24. Then I'd add those together. But something is making me pause---am I reading 'half as many' right? For a second, I start doubting myself and get caught up rereading the question. My brain's kind of spinning in circles now, so I think I need a break. I get up, pour myself some water, and just stand by the window watching the cars go by. \ldots \\
    \midrule
    Symmetry & When analyzing a problem, look for elements or relationships that remain unchanged when you alter your point of view, reorder components, or exchange similar parts. These invariants often reveal hidden simplicity and reduce the number of distinct cases you must consider. \ldots & Let me look for something that doesn't change even if I shuffle the information. Notice that it doesn't matter whether we talk about April first or May first---the relationship is always that May is half of April. If I imagine swapping their names or even combining both months into one big 'sales event,' the math stays the same. April's 48 clips and May's 24 clips always add up to 72, regardless of the order. \ldots \\
    \midrule
    Terms Old And New & As you solve the problem, explicitly focus on the words, labels, and symbols used in the problem statement---they shape your understanding. Begin by restating each key term, quantity, or symbol in your own words to ensure clarity and prevent ambiguity. \ldots & Let me make sure I understand what the problem is asking. Natalia sold clips in two different months: April and May. The statement says she sold clips to 48 of her friends in April. I'm going to take that to mean she sold 48 clips in April, one to each friend. Then it says that in May, she sold half as many clips as she did in April. So, for May, I just need to take half of 48. Half of 48 is 24. \ldots \\
    \midrule
    Test By Dimension & Whenever you form an equation, relationship, or comparison, check that the quantities or ideas involved are of the same kind. Ask yourself: can these things meaningfully be added, compared, or equated? Verify that their units, scales, or logical categories match --- length with length, rate with rate, cost with cost, cause with effect. \ldots & Natalia sold 48 clips in April. I notice that this '48' is counting the number of clips, not dollars or people or any other category. In May, she sold half as many clips. Since 'half as many' must refer to the same kind of thing---clips---I'm safe to divide 48 by 2, which gives 24 clips for May. Now, before adding, I make sure both numbers are counts of clips---48 from April, 24 from May. Since they match, I add them directly: 48 plus 24 equals 72 clips sold altogether. \\
    \midrule
    The Future Mathematician & Approach every problem with the curiosity of someone who wants to understand, not just to answer. Be patient with difficulty --- explore, test, and learn from each attempt rather than rushing to a result. Reflect on your reasoning: ask yourself why each step works and what it reveals. When you find an idea that helps, remember it; when you make a mistake, study it. \ldots & Before jumping into numbers, I want to really understand what this problem is saying. Natalia sold 48 clips to her friends in April. Then in May, she sold half as many. I'm careful with phrases like 'half as many'---does that mean I take 48 and divide it by 2? Let me check: 48 divided by 2 is 24, so May's total should be 24. Now, to find the total clips for April and May, I add 48 and 24, which gives 72. But I don't want to stop there. \ldots \\
    \midrule
    The Intelligent Problem Solver & As an intelligent problem-solver, approach each problem actively and independently: Select your own methods and strategies, rather than relying solely on prescribed formulas. Justify each step and decision, explaining why you choose one approach over another. Adapt your plan if you see a more effective path, and take responsibility for carrying your solution through, checking your work as you go. \ldots & Let me approach this problem the way an independent problem-solver would, not just plugging in numbers but thinking through my choices. Natalia sold 48 clips in April. The next piece says she sold half as many in May. Before calculating, I'll ask myself: does 'half as many' mean I should take 48 and divide by 2, or could it mean something else? I decide dividing by 2 is most natural here, giving me 24 for May. \ldots \\
    \midrule
    The Intelligent Reader & When reading a mathematical solution or argument, do not simply accept each step as correct. Actively engage with the text: ask yourself what each step accomplishes, why it is needed, and how it connects to the overall purpose. Pause at every new statement or operation and try to see not only that it is valid, but also why it is made---what motivated it, and how it advances the solution. \ldots & Let me not just accept the numbers at face value---what is the first statement doing for me? It says Natalia sold clips to 48 friends in April. Is that just a fact to note, or does it set the stage for something? The next line says she sold half as many clips in May. Wait, why is May described as 'half as many'? That tells me May isn't independent; it's built on April's number. \ldots \\
    \midrule
    The Traditional Math Prof & First, solve the problem by rigidly following formal rules and procedures, focusing on definitions and formulas without seeking deeper understanding. As you work, notice that this approach feels mechanical or unsatisfying. After noticing this, pause, acknowledge the limitation, and restart---this time focusing on understanding, intuition, and the reasons behind each step. \ldots & Let us proceed as is customary: In April, the problem states 48 clips were sold. For May, 'half as many' means 48 divided by 2, which yields 24. Adding the months, 48 plus 24 gives 72. That, by the established method, is the answer. Now, I should be satisfied, but as a professor, I confess that---despite the logical correctness---such a solution leaves me oddly unfulfilled. The process, for all its rigor, neglects the spirit behind the calculation, the connection between the months. \ldots \\
    \midrule
    Variation Of Problem & When you solve a problem, you will need a deeper understanding. For that, change the problem itself. Try varying its data, goal, or condition --- make it simpler, broader, or slightly different. Ask: What happens if one assumption changes? If I look for a different unknown? If I add, remove, or reverse a condition? Study how the solution changes --- or doesn't --- under these variations. \ldots & Okay, let me mess around with the numbers for a second. What if instead of 48, Natalia sold 60 clips in April? I just want to see what happens. So, half as many in May would be 30, right? That adds up to 90 clips. Now, if I go back to the original---48 in April, half as many in May is 24, and 48 + 24 is 72. Huh. Changing April's number just scales the total, but the relationship---April plus half of April---never changes. That makes the structure of the problem really clear. \\
    \midrule
    What Is The Unknown & Begin every problem by identifying exactly what you are asked to find or determine. Ask yourself: What is missing? What do I need to know that I don't yet know? Describe the unknown clearly in words or represent it with a symbol or variable. Think about its nature --- is it a number, a relationship, a pattern, a plan, or a proof? \ldots & Before I can begin solving, I need to ask: What am I trying to find here? The problem talks about Natalia selling clips over two months, April and May, but what is actually missing? The unknown is the total number of clips Natalia sold in both months combined. To clarify, I'll represent this unknown as T. My goal is to figure out the value of T, where T = (clips in April) + (clips in May). \ldots \\
    \midrule
    Why Proofs & After you find an answer or discover a pattern, always ask yourself whether it is necessarily true or just appears to be. Seek proof---not merely as a formality, but as a means to transform your intuition or evidence into certainty. A proper proof should convince both you and others, explaining exactly why the result must hold and under which conditions it applies. \ldots & At first glance, it seems straightforward: Natalia sold 48 clips in April and half as many in May. But instead of just accepting this, I need to prove---beyond intuition---that the total is necessarily 72. Let's check the logic. If April is 48, then by definition, 'half as many' in May is 48 divided by 2, giving 24. Now, adding these, 48 + 24 = 72. \ldots \\
    \midrule
    Wisdom Of Proverbs & When reasoning or solving problems, draw upon the timeless insights encoded in old proverbs --- each expressing deep heuristic wisdom in simple language. Begin your reasoning by quoting a relevant proverb. Then, interpret the proverb as a guiding principle for the problem at hand: use it to shape your reasoning, attitude, and approach --- to think carefully before acting, persist through difficult \ldots & Natalia sold 48 clips in April. Now, for May, I notice she sold half as many, which means 48 divided by 2 --- that's 24. At this point, I'm reminded, 'Measure twice, cut once.' Before I add, let me double-check that math: half of 48 is indeed 24. Now, adding both months together, 48 plus 24 gives 72. Careful checking helps avoid mistakes, just like the proverb says. \\
    \midrule
    Working Backwards & You must start from the goal itself. Ask what must be true immediately before the goal could be achieved, and continue reasoning backward step by step until you reach a known fact or the given data. Once you have traced this reverse path, follow it forward again to confirm the solution. \ldots & Let me approach this problem by starting from the end---what I'm really after is the total number of clips Natalia sold over April and May. To get there, I need to think about what must be true right before I have that total. The sum must come from the individual sales in April and in May. So, to reconstruct the total, I need to know how many clips were sold in each month. I know April's number is given directly as 48. But I don't know May yet. \ldots \\
    \bottomrule
\label{tab:heuristics-complete}
\end{longtable*}
\normalsize
\setlength{\tabcolsep}{6pt}
\setlength{\baselineskip}{\normalbaselineskip}
\renewcommand{\arraystretch}{1.0}





\end{document}